\documentclass[format=acmsmall,screen=true,nonacm]{acmart}

\usepackage{booktabs}
\graphicspath{{pdf}}
\DeclareGraphicsExtensions{.pdf,.png}
\usepackage{bigstrut}
\usepackage{multirow}
\usepackage{multicol}

\newcommand{\p}[1]{\smallskip \noindent \textbf{{#1}.}}
\newcommand{\eq}[1]{Equation~(\ref{eq:#1})}
\newcommand{\fig}[1]{Figure~\ref{fig:#1}}

\usepackage{graphicx}
\graphicspath{{pdf}}
\DeclareGraphicsExtensions{.pdf,.png}

\usepackage{csquotes}
\usepackage{balance}
\usepackage{amsmath}
\DeclareMathOperator*{\argmax}{arg\,max}
\usepackage{xcolor}
\usepackage{algorithm}
\usepackage{algpseudocode}

\setcopyright{none}

\begin{document}

\title{Unified Learning from Demonstrations, Corrections, and Preferences during Physical Human-Robot Interaction}

\author{Shaunak A. Mehta}
\orcid{0000-0002-3160-6879}
\email{mehtashaunak@vt.edu}

\author{Dylan P. Losey}
\orcid{0000-0002-8787-5293}
\email{losey@vt.edu}

\affiliation{
  \institution{Virginia Tech}
  \department{Department of Mechanical Engineering}
  \streetaddress{635 Prices Fork Rd}
  \city{Blacksburg}
  \state{VA}
  \postcode{24060}
  \country{USA}
  }

\begin{abstract}

Humans can leverage physical interaction to teach robot arms. This physical interaction takes multiple forms depending on the task, the user, and what the robot has learned so far. State-of-the-art approaches focus on learning from a single modality, or combine some interaction types. Some methods do so by assuming that the robot has prior information about the features of the task and the reward structure. By contrast, in this paper we introduce an algorithmic formalism that unites learning from demonstrations, corrections, and preferences. Our approach makes no assumptions about the tasks the human wants to teach the robot; instead, we learn a reward model from scratch by comparing the human’s input to nearby alternatives, i.e., trajectories close to the human’s feedback. We first derive a loss function that trains an ensemble of reward models to match the human's demonstrations, corrections, and preferences. The type and order of feedback is up to the human teacher: we enable the robot to collect this feedback passively or actively. We then apply constrained optimization to convert our learned reward into a desired robot trajectory. Through simulations and a user study we demonstrate that our proposed approach more accurately learns manipulation tasks from physical human interaction than existing baselines, particularly when the robot is faced with new or unexpected objectives. Videos of our user study are available at: \href{https://youtu.be/FSUJsTYvEKU}{https://youtu.be/FSUJsTYvEKU}

\end{abstract}

%
%



%
%



\maketitle

\section{Introduction}

\begin{figure}[t]
	\begin{center}
		\includegraphics[width=1.0\columnwidth]{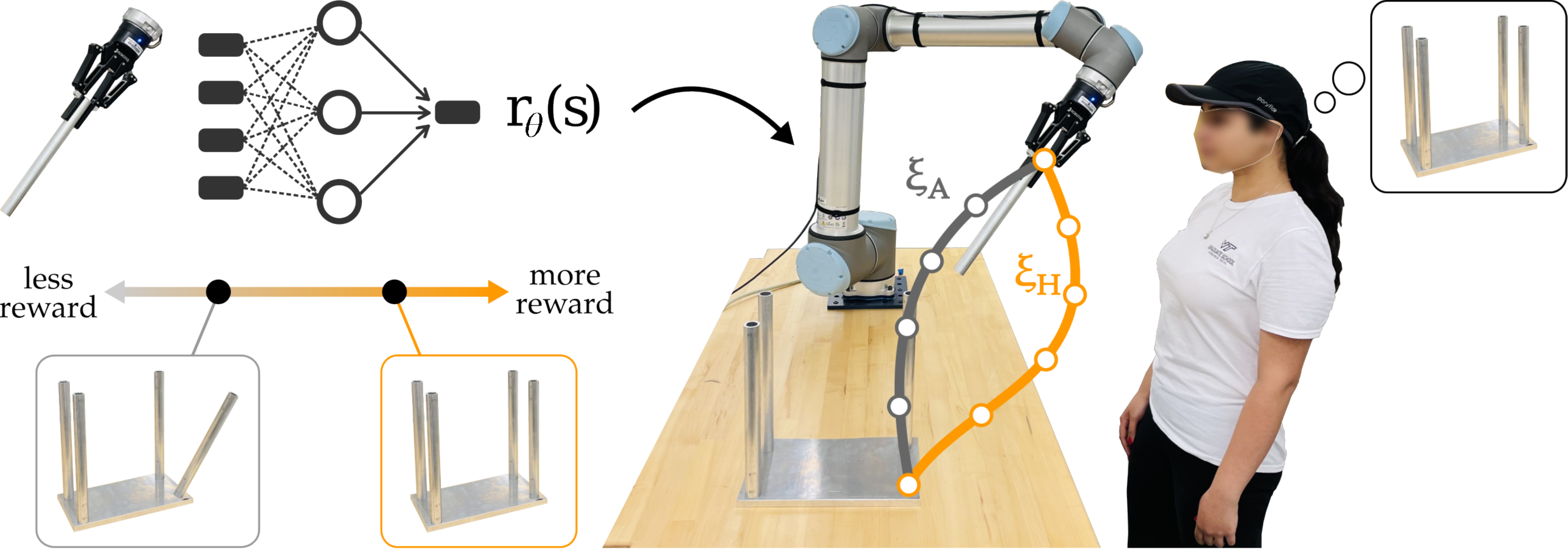}
		\caption{Human teaching a robot arm to assemble a chair. The robot does not have any prior information about this task, and must learn from the human's physical interactions. We recognize that these interactions can take multiple different forms, including demonstrations, corrections, and preferences. To unify each type of input under a single framework, we train a reward model to assign higher scores to the human's behavior ($\xi_R$) than to nearby alternatives ($\xi_A$). The robot then optimizes this reward model to find its desired trajectory.}
		\label{fig:front}
	\end{center}
\end{figure}

Imagine teaching a robot arm on a factory floor (see \fig{front}). You know what task you want the robot to perform --- attaching a chair leg --- but you do not know how to program the robot to perform this task. Instead, you \textit{physically} interact with the robot. Perhaps you start by kinesthetically guiding the robot across a full demonstration of the task. Next you deploy the robot, and notice it is attaching the chair leg at the wrong angle: here you physically intervene and correct just that part of the arm's motion. Finally, as the robot gets closer to understanding your task, you might rank the robot's behavior, indicating times where it gets it right and times where it messes up. Throughout this process it does not make sense for you to be constrained to only a single type of interaction (e.g., only ever showing demonstrations of the task). Instead, you should be able to exploit \textit{all} the avenues of physical human-robot interaction to convey your intent.

As robots increasingly share spaces with human partners, physical interaction between humans and robots becomes inevitable. Within this paper we focus on robot arms that perform manipulation tasks. Here we can harness physical interaction as a channel for communication; when the human kinesthetically guides a robot arm throughout their desired task, the human is providing a \textit{demonstration}; when the human pushes, pulls, or twists the robot to fix a part of its motion, the human is showing a \textit{correction}; and when the human physically observes the robot and ranks its performance, the human is indicating their \textit{preference}. Demonstrations, corrections, and preferences all communicate information about how the human wants the robot to behave. But we recognize that these modalities provide different --- and often complementary --- types of information: demonstrations provide an outline of the high-level task, corrections hone-in on fine-grained aspects of the robot's motion, and preferences provide a ranked comparison between various repetitions of the same task. The right type of interaction (e.g., demonstration, correction, or preference) depends on the task, the user, and what the robot has learned so far.

Recent research on physical human-robot interaction develops separate approaches for each type of human feedback. Robot arms can learn manipulation tasks only from demonstrations \cite{argall2009survey}, only from corrections \cite{losey2021physical}, or only from preferences \cite{jain2015learning}. Moving beyond physical human-robot interaction, there is also work that unites combinations of these data sources: for instance, learning from demonstrations and preferences \cite{ibarz2018reward, brown2019extrapolating, chen2021learning, zhang2021confidence, biyik2021learning}, or from demonstrations and corrections \cite{ross2011reduction, spencer2022expert, kelly2019hg, hoque2021thriftydagger}. But to learn from multiple sources of human feedback the robot must either (a) have prior information about the tasks the human has in mind \cite{biyik2021learning, jeon2020reward} or (b) apply reinforcement learning to identify the optimal trajectory \cite{ibarz2018reward, brown2019extrapolating, chen2021learning, zhang2021confidence, christiano2017deep}. These constraints present practical challenges during physical human-robot interaction: intuitively, we seek robots that learn new and unexpected tasks in real-time, without stopping to perform reinforcement learning through trial and error.

In this paper we propose a formalism for learning from physical interaction that unites demonstrations, corrections, and preferences. We recognize that these different types of interactions provide different types of data about the human's desired task. To ground each feedback type within the same learning framework, our hypothesis is that:
\begin{center}
\textit{We can unite physical demonstrations, corrections, and preferences under a single framework to learn a reward model that assigns higher scores to the human's input trajectories as compared to any modified alternatives.}
\end{center}
Applying this insight we develop a two-step algorithm. First, we observe the human's physical interactions with the robot and learn a \textit{reward model} from scratch. Second, we exploit the underlying structure of manipulation tasks by applying \textit{constrained optimization} to map the learned reward into a robot trajectory. This process is iterative and free-form: at every iteration the human can provide demonstrations, corrections, or preferences, and the robot updates its reward model and identifies an optimal trajectory in real-time. We emphasize that the reward model is a neural network that does not require any \textit{a priori} knowledge about the human's potential rewards, and thus the current user is able to teach the robot arbitrary manipulation tasks. Returning to our running example, imagine that the robot has never assembled a chair before: but because the robot learns from the human's feedback to assign higher rewards to states where the chair legs are upright, the robot solves for a desired trajectory that carries the legs vertically.

Overall, we make the following contributions to physical human-robot interaction:

\p{Uniting Physical Interactions} We present a learning approach to physical interaction that unifies demonstrations, corrections, and preferences. Our approach is based on the insight that the human's inputs are better than the alternatives: e.g., the human's demonstrated trajectory should receive higher reward than noisy perturbations of that same trajectory. Our learning approach automatically generates trajectory deformations of the human's physical interactions, and then learns to assign higher rewards to the human's actual behavior.

\p{A Flexible Reward Learning Framework} We incorporate learning from both active and passive sources of feedback into a flexible reward learning framework. In a setting where the human may use multiple forms of feedback to teach the robot, we also enable the robot to actively prompt the human by eliciting their preferences. We identify a method for generating preference trajectories that — when ranked by the human — minimizes the robot’s uncertainty over its learned reward. This approach results in an end-to-end model of the human’s reward function. We then harness off-the-shelf optimization techniques and the robot’s underlying kinematics to convert this learned reward function into a desired robot trajectory.

\p{Comparing to Baselines} We compare our approach to state-of-the-art baselines across experiments with real robot arms and simulated and real human users. First, we consider approaches for physical human-robot interaction that learn from either demonstrations and corrections or demonstrations and preferences, and show that our method more accurately learns the human's task (particularly when this task is new or unexpected). Second, we perform a user study with imitation learning baselines that synthesize multiple forms of human feedback. Here we show that participants prefer working with robots that learn using our approach, and that our approach learns trajectories that better align with the human's intended tasks. Readers can find videos of our experimental setup and user study at: \href{https://youtu.be/FSUJsTYvEKU}{https://youtu.be/FSUJsTYvEKU}
\section{Related Work}\label{sec:related}

This paper introduces a learning formalism for physical human-robot interaction. During physical interaction the human can kinesthetically guide the robot throughout examples of the task (demonstrations), apply forces and torques to adjust segments of the robot's motion (corrections), and rank the robot trajectories that they observe (preferences). Our approach seeks to unite demonstrations, corrections, and preferences into a real-time learning algorithm. To enable safe and seamless human-robot interaction, we first take advantage of prior research on shared control. We then review two related topics from the existing literature: (a) learning methods designed specifically for physical human-robot interaction and (b) approaches outside of physical interaction that learn a reward function from multiple sources of human feedback.

\p{Control Responses for Physical Interaction} Although we will focus on learning, we recognize that prior research has also explored control theoretic responses to physical interaction \cite{hogan1984impedance, de2008atlas, haddadin2016physical, losey2018review, haddadin2008collision, li2019differential, music2017control, mortl2012role}. Works on impedance control and shared control arbitrate leader-and-follower roles between the human and robot: when the human intervenes and applies large forces to the robot, the robot reduces its control feedback so that the human backdrives the arm. Intuitively, the robot can also pause, move away from the human, or follow the human's motion during physical interaction \cite{haddadin2008collision}. These control responses are an important step towards safety, and we will leverage shared control throughout this paper to enable close physical interaction.

\p{Learning Rewards during Physical Interaction} Recent work has enabled robots to learn from physical human-robot interaction \cite{argall2009survey}. Here the human physically pushes, pulls, and twists the robot arm, and the robot attempts to infer why the human is applying these forces so that it can update its behavior accordingly. Some works directly map the human's forces and torques to changes in the robot's desired trajectory \cite{akgun2012keyframe, losey2019learning, khoramshahi2019dynamical, hagenow2021corrective}. But more relevant here is research that infers a reward function from physical interaction: these approaches assume that the human has in mind an objective, and the human's interactions are observations of this latent reward function \cite{osa2018algorithmic, abbeel2004apprenticeship, ng2000algorithms}. The learned reward is then used to identify the robot's desired trajectory.

In practice, these approaches often take advantage of physical human corrections \cite{losey2021physical, bobu2022inducing, jain2015learning, kalakrishnan2013learning, yin2019ensemble, li2021learning}. Using the insight that the human's applied forces and torques are an intentional improvement --- i.e., the human is going out of their way to show the robot a better way to perform the task --- the robot learns to give the human's behavior higher rewards and propagate those changes the next time the robot repeats this task. In this paper we leverage a similar insight to derive our learning algorithm. However, while prior works for physical human-robot interaction focus on a single input modality (e.g., physical corrections), we will develop a framework that incorporates the different aspects of physical feedback.

\p{Learning Rewards from Multiple Types of Feedback} Outside of physical interaction several methods have been proposed to learn from different types of human feedback. For example, interactive imitation learning approaches can synthesize demonstrations and corrections \cite{ross2011reduction, spencer2022expert, kelly2019hg, hoque2021thriftydagger}. Consider a human watching a mobile robot: first the human might teleoperate the robot throughout several laps of the building, and then the human may jump in and correct the robot only when it makes a specific mistake. Alternatively, methods that learn from suboptimal humans often combine demonstrations and preferences \cite{ibarz2018reward, brown2019extrapolating, chen2021learning, zhang2021confidence, lee2021pebble, christiano2017deep}. Imagine that you are teaching a simulated robot to play an Atari game. After you do your best to provide a demonstration --- and score as high as possible --- you can rank the robot's autonomous performance to indicate when it is performing well and when it is making mistakes. Most relevant to our research are \cite{jeon2020reward} and \cite{biyik2021learning}, where the authors unite different types of human feedback to learn a single reward model. 

When applying these existing approaches to physical human-robot interaction we are faced with two problems. On the one hand, methods like \cite{jeon2020reward, biyik2021learning} require prior knowledge about the human's reward function --- as we will show in our analysis and experiments, these methods fall short when the human wants to teach the robot a new or unexpected task. On the other hand, approaches like \cite{ibarz2018reward, brown2019extrapolating, chen2021learning, zhang2021confidence, lee2021pebble, christiano2017deep} use reinforcement learning to convert the reward function into robot behavior. But reinforcement learning is time consuming and requires trial and error --- which may not be possible (or safe) when humans are physically interacting with robot arms. Accordingly, in this paper we introduce a formalism that unites demonstrations, corrections, and preferences without the need for pre-defined tasks or reinforcement learning.

\section{Problem Statement} \label{sec:problem}

Going back to our motivating example from \fig{front}, the human wants to teach their robot arm how to perform a manipulation task. To teach the robot the human exploits physical interaction: the human kinesthetically guides the robot through the process of attaching a chair leg, modifies specific sections of the robot's trajectory, and ranks the robot's autonomous behavior across repeated interactions. In this section we formulate the problem of learning from each of these different types of physical interaction. We explain how existing approaches combine demonstrations, corrections, and preferences when the robot has prior knowledge about the tasks it will perform, and then highlight the shortcomings of these assumptions. Finally, we define our proposed reward model: we highlight that this approach can take advantage of the assumptions from previous works, but is also capable of learning new tasks that the robot does not know about beforehand.

\p{Task} We formulate the task that the human wants the robot to perform as a Markov decision process: $\mathcal{M} = \langle \mathcal{S, A}, T, r, H \rangle$. Here $s \in \mathcal{S}$ is the state of the robot and $a \in \mathcal{A}$ denotes the robot action. For instance, in our motivating example the robot's state $s$ consists of its joint position and the position of objects in the scene, and the robot's action $a$ is its joint velocity (e.g., a change in joint position). At timestep $t$ the robot transitions to a new state based on the dynamics $s_{t+1} = T(s_t, a_t)$. The task ends after a total of $H$ timesteps.

Remember that --- although the human knows what task the robot should perform --- the robot may be uncertain about the correct task. We capture this objective using the reward function $r : \mathcal{S} \rightarrow \mathbb{R}$. The reward function maps robot states to scalar values (where higher scores indicate better states). To complete the task successfully the robot must maximize the reward function; we will enable the robot to learn this reward function from physical interaction.

\p{Trajectory} During each iteration of the task the robot moves through a sequence of $H$ states. We refer to this sequence as a trajectory $\xi \in \Xi$ such that $\xi = (s_0, s_1, \ldots s_H)$.

\begin{figure}
    \centering
    \includegraphics[width=1\columnwidth]{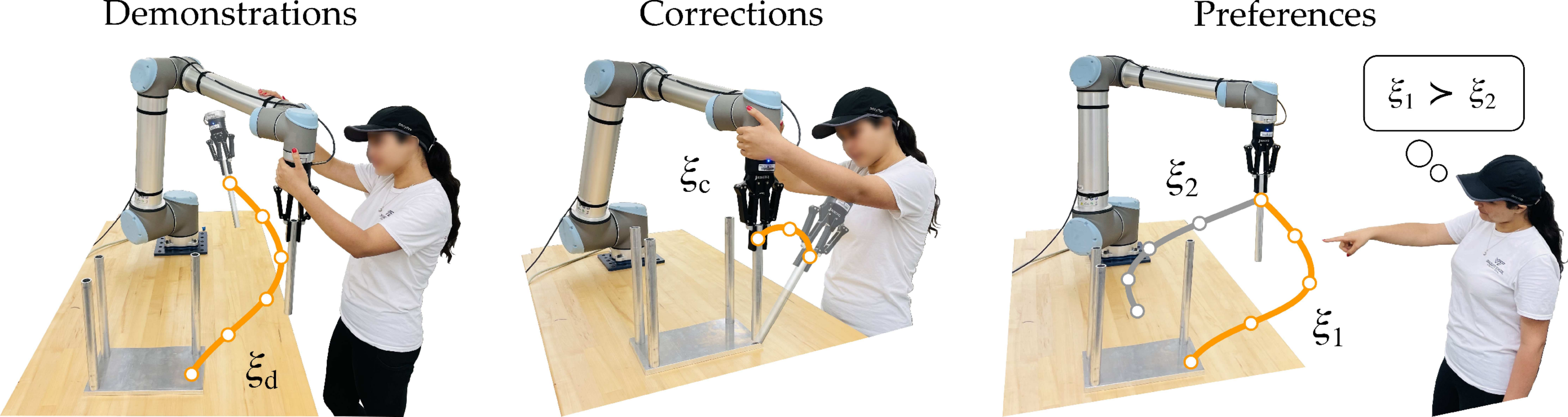}
    \caption{Different types of physical feedback. (Left) Humans can convey information to robot arms by kinesthetically guiding the robot through a demonstration of the task. Demonstrations provide high-level information about the entire trajectory. (Middle) To refine a specific part of the robot's motion humans may make physical corrections. These corrections fine-tune the robot's behavior. (Right) Over repeated interactions the human will observe multiple robot trajectories. Humans can rank these trajectories (i.e., give their preferences) to indicate when the robot is making a mistake. We note that preferences are not physical --- in the sense that the human does not apply forces or torques --- but preference feedback naturally emerges when humans and robots occupy the same space and the human can physically observe the robot's behavior.}
    \label{fig:types}
\end{figure}

\p{Demonstrations} The first way that the human can physically communicate with the robot arm is by providing demonstrations (see~\fig{types}). During a demonstration the human kinesthetically backdrives the robot throughout a trajectory $\xi_d$. We refer to the set of demonstrations as $\mathcal{D} = \{\xi_{d_1}, \xi_{d_2}, \ldots\}$, where each element of $\mathcal{D}$ is an $H$-length trajectory that shows the entire task.

\p{Corrections} During demonstrations the robot is passive throughout the entire interaction. But when the robot attempts to perform the task autonomously, the human may intervene to correct just a snippet of the robot's motion. Let $\xi_r \in \Xi$ be the trajectory that the robot is autonomously executing, and let $\xi_c \in \Xi_{\mathcal{C}} \subset \Xi$ be the human's correction. We emphasize that $\xi_c$ includes fewer than $H$ timesteps, and the human only physically intervenes to push, pull, and guide the robot during $\xi_c$. Let the set of human corrections be written as $\mathcal{C} = \{(\xi_{r_1}, \xi_{c_1}), (\xi_{r_2}, \xi_{c_2}), \ldots \}$. Each correction consists of both the robot's initial trajectory and the human's modification.

\p{Preferences} For demonstrations and corrections the human is physically in contact with the robot arm. But the human and robot are also sharing the same space: and thus the human can observe the robot's trajectories and provide feedback about its behavior. Hence, the final way that the human conveys information to the robot is through preferences. We define a preference query as a set of $k$ trajectories ranked in order by the human: $Q = \{ \xi_{1} \succ \xi_{2} \succ, \ldots, \succ \xi_{k}\}$. Put another way, the human thinks $\xi_{1}$ better aligns with the task's reward than $\xi_{2}$. We group the human's preferences into a set $\mathcal{Q} = \{Q_1, Q_2, \ldots \}$, where each element of $\mathcal{Q}$ is a set of $k$ ranked trajectories.

\subsection{Preliminaries: Learning Rewards with Known Features}\label{subsec:prelim}

In order to formulate our problem we first need to explain how other methods learn rewards from demonstrations, corrections, and preferences. Remember that we defined the reward function $r$ as an arbitrary mapping from states to scores. Related works such as \cite{osa2018algorithmic, jeon2020reward, losey2021physical, biyik2021learning, ziebart2008maximum} introduce structural bias by assuming that the reward function is a linear combination of \textit{features}:
\begin{equation} \label{eq:P1}
    r_{\theta}(s) = \theta \cdot \phi(s)
\end{equation}
Here $\phi : \mathcal{S} \rightarrow \mathbb{R}^n$ is the feature vector and $\theta \in \mathbb{R}^n$ is a weight vector. Features capture metrics that are potentially relevant to the current task. Within our running example from \fig{front}, features could include the robot's distance from the chair, the angle of the chair leg, and the orientation of the chair. The robot is given these features \textit{a priori} and must determine which features are important to the human; i.e., the robot is given $\phi$ and must learn $\theta$. Sticking with our running example, the robot should learn to assign higher weight to the angle of the leg and lower weight to the orientation of the chair.

If we assume that the robot has access to the task-relevant features --- and thus the reward is structured as in \eq{P1} --- we can use Bayesian inference to learn the correct weights $\theta$. Let $P(\theta \mid \mathcal{D}, \mathcal{C}, \mathcal{Q})$ be the probability of weights $\theta$ given the human's previous demonstrations, corrections, and preferences. Applying Bayes' rule, and recognizing that each type of feedback is conditionally independent, we reach:
\begin{equation} \label{eq:P2}
    P(\theta \mid \mathcal{D}, \mathcal{C}, \mathcal{Q}) \propto P(\mathcal{D} \mid \theta) \cdot P(\mathcal{C} \mid \theta) \cdot P(\mathcal{Q} \mid \theta) \cdot P(\theta)
\end{equation}
Here $P(\theta)$ is the prior distribution over $\theta$, and the $P( \, \cdot \mid \theta)$ terms capture the likelihood of the observed demonstrations, corrections, or preferences given that the human has reward weights $\theta$. Previous works have found expressions for these likelihood functions \cite{jeon2020reward}. For instance, we can model humans as approximately optimal, so that human inputs with higher rewards are increasingly likely \cite{ziebart2008maximum}: $P(\xi_d \mid \theta) \propto \exp(\sum_{s \in \xi_d} \theta \cdot \phi(s))$. What is important here is that --- if we assume access to the task-relevant features --- inferring $\theta$ simplifies to \eq{P2}. 

This preliminary approach makes sense if robots have access to all their reward features \textit{a priori}. In practice, however, robots will inevitably face tasks they did not expect and features that were not pre-programmed \cite{bobu2022inducing}. Consider our motivating example: when we first bring this robot arm onto the factory floor, will the robot understand the features of chair orientation or leg attachments? Human users should not be forced to hand-engineer features for each new task and environment; when the features are available, the robot arm should make use of their structure --- but when the human's physical interactions are not aligned with any known feature, the robot should not be constrained to misspecified feature spaces \cite{bobu2020quantifying}. Instead, we will develop robots that can learn reward functions \textit{without} requiring predefined features.

\subsection{Problem: Learning Arbitrary Rewards from Physical Interaction}\label{subsec:problem}

Our goal is (a) to learn the task reward function from demonstrations, corrections, and preferences and then (b) to leverage this learned reward to identify an optimal robot trajectory that performs the task autonomously. To remove the reliance on features --- and enable the robot to learn arbitrary task rewards --- we will model the reward function as a neural network:
\begin{equation} \label{eq:P3}
    R_{\theta}(\xi) = \sum_{s \in \xi} r_{\theta}(s)
\end{equation}
Here $\theta$ is the weights of the neural network, $r_{\theta}(s)$ is the learned reward at state $s$, and $R_{\theta}(\xi)$ is the cumulative reward along trajectory $\xi$. If features are available we can incorporate them within this formulation. Define $s = \big(s, \phi(s)\big)$ as an augmented state vector which now includes both the system state and the features $\phi(s)$. Returning to \eq{P3}, we learn a reward model $r_{\theta}$ that maps this augmented state to reward values: cases where the task reward simplifies to $\theta \cdot \phi(s)$ are a special instance of our more general formulation. 

We have chosen to learn a reward function because it provides an avenue to unify demonstrations, corrections, and preferences. In the next section we will develop an algorithm to train this reward function from each different type of physical interaction.

\section{Unifying Demonstrations, Corrections, and Preferences} \label{sec:method}

Our learning approach is based on comparisons. Recall that our underlying hypothesis is that the human's inputs --- whether they are demonstrations, corrections, or preferences --- are \textit{intentional} improvements to the robot's behavior. Put another way, the reward model in \eq{P3} should assign higher scores to human trajectories than to nearby alternatives. In this section we apply our insight to develop a unified learning algorithm. First, we explain how to generate trajectory deformations that we can compare against the human's inputs. Next, we train the reward function to score the human's demonstrations, corrections, or preferences higher than these noisy alternatives. Finally, we leverage constrained optimization to convert our learned reward model into a robot trajectory. Throughout this section we consider both passive communication (where the human chooses how to physically intervene) and active information gathering (where the robot prompts the human to uncover the correct reward function). We emphasize that our resulting approach is flexible, and humans can teach the robot using whichever physical feedback modalities they prefer.

\subsection{Learning the Reward Model}

\p{Generating Trajectories for Comparison} Given an input trajectory $\xi$ we first seek to generate a nearby alternative $\hat{\xi}$. The intuition here is that the human chose to input $\xi$ and not $\hat{\xi}$ --- and thus the robot should assign higher rewards to $\xi$ as compared to $\hat{\xi}$.

\begin{figure}
    \centering
    \includegraphics[width=1.0\columnwidth]{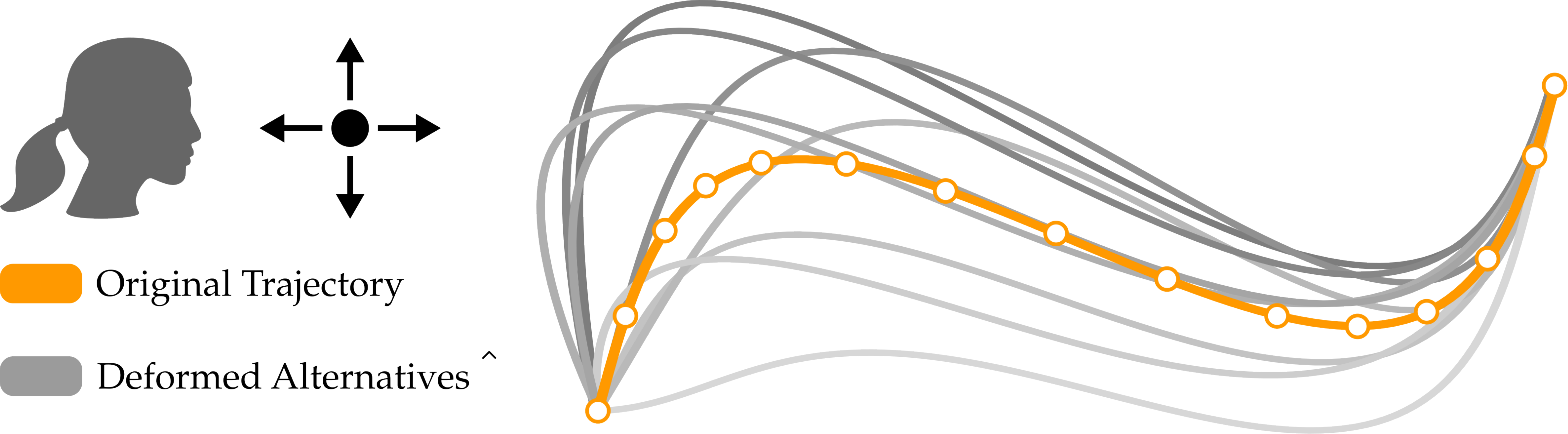}
    \caption{Generating trajectories for comparison. In this example the human moves a $2$-DoF point mass robot along a sine wave. We record the initial trajectory $\xi$, and then apply \eq{M1} to generate smooth perturbations $\hat{\xi}$. Our learned reward model should score $\xi$ as a better trajectory than any of the alternatives $\hat{\xi}$.}
    \label{fig:deformations}
\end{figure}

To create alternatives we propagate noisy perturbations along the input trajectory following the approach outlined in \cite{dragan2015movement, losey2017trajectory}:
\begin{equation} \label{eq:M1}
    \hat{\xi} = \xi + M \lambda
\end{equation}
Here $\lambda \in \mathbb{R}^H$ is a noise vector that the designer selects and $M \in \mathbb{R}^{H \times H}$ is a symmetric positive definite matrix that defines a norm on the trajectory space. Our approach is not tied to any specific choice of $M$ or $\lambda$; however, for our experiments we selected the acceleration norm \cite{dragan2015movement, losey2017trajectory} in order to get \textit{smooth} trajectory deformations of $\xi$:
\begin{equation} \label{eq:M2}
    M = (A^T A)^{-1}, \quad A =\begin{bmatrix}
    \begin{matrix}1 & 0 & 0 \\ -2 & 1 & 0 \\ 1 & -2 & 1 \\ 0 & 1 & -2 \\ 0 & 0 & 1\end{matrix} 
     & \cdots & 
     \begin{matrix}0 \\ 0\\ 0\\ 0\\ 0\end{matrix}\\
     \vdots & \ddots & \vdots\\
     \begin{matrix}0 & 0 & 0\\ 0 & 0 & 0 \\ 0 & 0 & 0\end{matrix} 
     & \cdots &
     \begin{matrix}1 \\ -2 \\ 1 \end{matrix}
    \end{bmatrix}
\end{equation}
Each deformation uses the same $M$ and a new vector $\lambda$. In our experiments we sampled $\lambda$ from a zero-mean Gaussian distribution, and for every sampled $\lambda$ we generated the corresponding $\hat{\xi}$ --- to visualize this process, we show an example $\xi$ and the generated alternatives $\hat{\xi}$ in \fig{deformations}.

\p{Learning from Demonstrations} Now that we have a way to generate nearby trajectories, we will learn the reward function by comparing these alternatives to the human's inputs. We start with demonstrations. Assuming that the human provides near-optimal demonstrations, the robot should learn to match each demonstration $\xi_d \in \mathcal{D}$. More formally, the robot should learn a reward model such that $R\big(\xi_d\big) > R\big(\hat{\xi}_d\big)$, where $\hat{\xi}_d$ is a deformation found using \eq{M1}.

Let $P(x \succ y \mid \theta)$ be the likelihood --- from the robot's perspective --- that trajectory $x$ has higher total reward than trajectory $y$ given that the robot has learned reward weights $\theta$. Inspired by prior work on human decision making and Luce's choice axiom \cite{luce2012individual, lucas2008rational, shepard1957stimulus}, we write this probability as a softmax-normalized distribution:
\begin{equation} \label{eq:M3}
    P\big(\xi_d \succ \hat{\xi}_d \mid \theta \big) = \frac{\exp R_{\theta}\big(\xi_d\big)}{\exp R_{\theta}\big(\xi_d\big) + \exp R_{\theta}\big(\hat{\xi}_d\big)}
\end{equation}
Remember that we want the robot to assign higher rewards to $\xi_d$ as compared to $\hat{\xi}_d$. When this happens we have that $P\big(\xi_d \succ \hat{\xi}_d \mid \theta \big) \rightarrow 1$ in \eq{M3}. Accordingly, to drive the probability $P\big(\xi_d \succ \hat{\xi}_d \mid \theta \big) \rightarrow 1$, we train the reward model to minimize the cross entropy loss \cite{ibarz2018reward, christiano2017deep, brown2019extrapolating, bobu2022inducing}:
\begin{equation} \label{eq:M4}
    \mathcal{L}_{\mathcal{D}}(\theta) = -\sum_{\xi_d \in \mathcal{D}} \mathbb{E}_{\hat{\xi}_d \sim \xi_d} \Big[ \log P\big(\xi_d \succ \hat{\xi}_d \mid \theta \big) \Big]
\end{equation}
Note that in \eq{M4} we sum the cross entropy loss across every demonstration $\xi_d \in \mathcal{D}$, and for each demonstration we sample a set of nearby trajectories $\hat{\xi}_d$. Reward models that minimize \eq{M4} will assign higher scores to trajectories that are like the human's demonstrations, and lower scores to trajectories that are different from these demonstrations.

\p{Learning from Corrections} During demonstrations the human backdrives the robot throughout the entire task; but during corrections, the human only fixes a snippet of the robot's trajectory. Given the robot's initial trajectory and the human's correction of this snippet --- i.e., $(\xi_{r}, \xi_{c}) \in \mathcal{C}$ --- we recognize that $R\big(\xi_c\big) > R\big(\hat{\xi}_r\big)$, where $\hat{\xi}_r$ is the segment of the robot's trajectory that the human has intentionally modified. More generally, we assume that the human's correction shows the robot the right way to perform this part of the task. We therefore have $R\big(\xi_c\big) > R\big(\hat{\xi}_c\big)$, where $\hat{\xi}_c$ is a deformation of just the human's correction. Following the same derivation that we applied for demonstrations, we reach the following loss function for corrections:
\begin{equation} \label{eq:M5}
    \mathcal{L}_{\mathcal{C}}(\theta) = \sum_{(\xi_r, \xi_c) \in \mathcal{C}} -\log \frac{\exp R_{\theta}\big(\xi_c\big)}{\exp R_{\theta}\big(\xi_c\big) + \exp R_{\theta}\big(\hat{\xi}_r\big)} - \mathbb{E}_{\hat{\xi}_c \sim \xi_c} \Bigg[ \log \frac{\exp R_{\theta}\big(\xi_c\big)}{\exp R_{\theta}\big(\xi_c\big) + \exp R_{\theta}\big(\hat{\xi}_c\big)} \Bigg] 
\end{equation}
Minimizing \eq{M5} encourages the robot to learn a reward function that classifies $\xi_c$ as a better trajectory than both the original segment $\hat{\xi}_r$ and perturbations $\hat{\xi}_c$ of the human's correction.

\p{Learning from Preferences} As a final form of feedback the human operator can rank the robot's physical trajectories. Consider the robot arm in \fig{front}: each time the robot attempts to add a chair leg, the human onlooker might score the robot's performance, and mark whether trajectory $\xi_i$ is better or worse than trajectory $\xi_j$. We emphasize that these preferences provide a \textit{direct} comparison between pairs of trajectories. Here we cannot assume that the human's preference is better than any other alternative; since the human's feedback only indicates that $\xi_i \succ \xi_j$, the reward model should learn $R\big(\xi_i\big) > R\big(\xi_j\big)$. Summing across the preferences $Q \in \mathcal{Q}$, we obtain the loss function:
\begin{equation} \label{eq:M6}
    \mathcal{L}_{\mathcal{Q}}(\theta) = -\sum_{Q \in \mathcal{Q}} \, \sum_{(\xi_i \succ \xi_j) \in Q} \log \frac{\exp R_{\theta}\big(\xi_i\big)}{\exp R_{\theta}\big(\xi_i\big) + \exp R_{\theta}\big(\xi_j\big)}
\end{equation}
where $(\xi_i \succ \xi_j)$ is any pair of human-ranked trajectories from preference $Q$. Intuitively, \eq{M6} trains the reward model to rank the robot's trajectories in the same order as the human operator. 

The human may passively provide these rankings over the course of repeated interactions. Alternatively, the robot can \textit{actively} prompt the human by rolling out a set of trajectories and asking for the human's preference. The goal of these active prompts is to reduce the robot's uncertainty about the correct task reward. To formalize this uncertainty we train an ensemble of $m$ rewards with weights $\mathcal{E} = \{\theta_1, \theta_2, \ldots, \theta_m \}$. Each of these reward models is a separate instantiation of \eq{P3}, and is trained using the same demonstrations, corrections, and preferences. Let $x$ and $y$ be two robot trajectories. If all reward models agree on the relative scores of these trajectories --- e.g., if $R_{\theta_i}(x) > R_{\theta_i}(y)$ for each $\theta_i \in \mathcal{E}$ --- then the robot is confident that $x \succ y$. But in cases where the reward models disagree --- for example, if $R_{\theta_1}(x) > R_{\theta_1}(y)$ while $R_{\theta_2}(x) < R_{\theta_2}(y)$ --- then we do not know which trajectory is better at the task. At times when the reward models \textit{disagree} the robot needs additional human feedback to resolve its uncertainty.

To actively learn about $\theta$ we will query the human by physically showing two robot trajectories, $\xi_1$ and $\xi_2$, and asking the human to choose which option they prefer. Humans may indicate that $\xi_1 \succ \xi_2$ or $\xi_2 \succ \xi_1$. Remember that we do not know beforehand how the human will respond to the robot; accordingly, we select $\xi_1$ and $\xi_2$ such that --- no matter which option the human chooses --- the robot maximizes the information it gains about the unknown task reward $\theta$:
\begin{equation} \label{eq:M7}
    (\xi_1^*, \xi_2^*) = \argmax_{\xi_1, \xi_2 \in \Xi} ~ \mathcal{I}\big(\theta ; \xi_i \mid (\xi_1, \xi_2)\big)
\end{equation}
Here $(\xi_1^*, \xi_2^*)$ is the greedily optimal query, $\xi_i$ is the human's preferred trajectory, and $\mathcal{I}$ is the information gain \cite{cover1999elements}. Following the derivation in \cite{biyik2021learning}, we find that the expected information gain from query $(\xi_1, \xi_2)$ across the ensemble $\mathcal{E}$ of reward models becomes:
\begin{equation} \label{eq:M8}
    \mathcal{I}\big(\theta ; \xi_i \mid (\xi_1, \xi_2)\big) = \frac{1}{m} \sum_{\xi_i \in Q} \sum_{\theta \in \mathcal{E}} P\big(\xi_i \succ \xi_{-i} \mid \theta\big) \log_2 \Bigg(\frac{m\cdot P\big(\xi_i \succ \xi_{-i} \mid \theta\big)}{\sum_{\theta' \in \mathcal{E}} P\big(\xi_i \succ \xi_{-i} \mid \theta'\big)}\Bigg)
\end{equation}
where $\xi_i$ is the trajectory the human prefers, $\xi_{-i}$ is the other trajectory, and $P\big(\xi_i \succ \xi_{-i} \mid \theta\big)$ is the softmax-normalized distribution from \eq{M3}. In practice, \eq{M8} looks for a query where (a) each reward model in the ensemble is confident about the relative scores of $\xi_1$ and $\xi_2$, but (b) some reward models think that $\xi_1 \succ \xi_2$, while other reward models think $\xi_2 \succ \xi_1$. We note that this active learning step is entirely optional. The robot still uses \eq{M6} to learn from human preferences regardless of whether they are obtained passively or actively. However, as we will show in our experiments, actively selecting prompts using \eq{M7} accelerates the robot's reward learning and resolves uncertainty across the ensemble of reward models.

\p{Putting It All Together} We have identified loss functions that train the reward model to match the human's demonstrations, corrections, and preferences. For each of these interaction modalities we have a common theme: the human's inputs should be scored higher than the alternatives. Our final step is to unite \eq{M4}, \eq{M5} and \eq{M6} into a single loss function:
\begin{equation} \label{eq:M9}
    \mathcal{L}(\theta) = \mathcal{L}_{\mathcal{D}}(\theta) + \mathcal{L}_{\mathcal{C}}(\theta) + \mathcal{L}_{\mathcal{Q}}(\theta)
\end{equation}
We note that \eq{M9} is our parallel to the Bayesian inference from \eq{P2}.

\floatname{algorithm}{Algorithm}
\begin{algorithm}[t]
\caption{Learning from Multiple Forms of Feedback}\label{alg:a1}
\begin{algorithmic}[1]
\State Randomly initialize the ensemble of reward models with weights $\mathcal{E}^i = \{\theta_0^i,\theta_1^i, \cdots \theta_m^i\}$
\State Initialize the Demonstration, Correction and Preference buffers $\mathcal{D}, \mathcal{C}, \mathcal{Q}$
\State Initialize the number of noisy alternatives for $\mathcal{D}$, $\mathcal{C}$ and $\mathcal{Q}$: $N_d, N_c, N_q$
\For {$i = 0, 1, 2, \cdots$}
    \State Initialize the rankings buffer $\mathcal{B}$
    \If {$ i = 0$ \textbf{and} Demonstration Provided}
        \State $\mathcal{D} \gets \xi_d$
    \ElsIf {Correction Provided}
        \State $\mathcal{C} \gets \xi_c$
    \ElsIf {Preference Provided}
        \State $\mathcal{Q} \gets Q$
    \EndIf
    \For {$j = 1, 2, \cdots N_d$}
        \State Generate noisy alternative $\hat{\xi}_d^j$ for $\xi_d \in \mathcal{D}$ \Comment{see eq. \ref{eq:M1}}
        \State $\mathcal{B} \gets (\xi_d^j \succ \hat{\xi}_d)$
    \EndFor
    \For {$j = 1, 2, \cdots N_c$}
        \State Generate noisy alternatives $\hat{\xi}_c^j$ for $\xi_c \in \mathcal{C}$ \Comment{see eq. \ref{eq:M1}}
        \State $\mathcal{B} \gets (\xi_c^j \succ \hat{\xi}_c)$
    \EndFor
    \For {$j = 1, 2, \cdots N_q$}
        \State Sample a preference $q^j \in \mathcal {Q}$
        \State $\mathcal{B} \gets q^j$
        \EndFor
    \State Update the reward models $\mathcal{E}^i$
\EndFor
\end{algorithmic}
\end{algorithm}

We outline the implementation procedure for our approach in Algorithm \ref{alg:a1}. By controlling the number of comparisons for each feedback type ($N_d$, $N_c$ and $N_p$), we can adjust their relative weight. Within our experiments we assign an equal importance to each of the feedback types. 
For our approach, the users can choose to provide any form of feedback to the robot by indicating their choice on a user interface. Previous work has proved that it is optimal to start with passive forms of feedback (e.g. demonstrations) before collecting active feedback (see Theorem 2 in  \cite{biyik2021learning}). Following this, the default order for our simulations and user study is demonstrations, then corrections, followed by preferences.

Given the human's inputs, we train an ensemble of reward models that minimize $\mathcal{L}(\theta)$. Each reward model is a fully connected network with two hidden layers and leaky rectified linear activation units. The output of the reward model is bounded between $-1$ and $+1$ using $\tanh(\cdot)$. Our ensemble includes three independently trained reward models: each model optimizes its weights using the Adam learning rule with an initial learning rate of $0.001$ \cite{kingma2014adam}. To compute the reward of a state $s$ we take the average score from $r_{\theta_1}(s)$, $r_{\theta_2}(s)$, and $r_{\theta_3}(s)$. We retrain the reward models after each new demonstration, correction, or preference from the human.

\subsection{Optimizing for Robot Trajectories}

The first half of our formalism is learning a reward function (or ensemble of reward functions) from the human's physical interactions. In the second part of our approach we convert this reward model $r_{\theta}$ into a robot trajectory $\xi_r$. Related approaches use reinforcement learning to identify the trajectory that maximizes $r_{\theta}$ \cite{ibarz2018reward, brown2019extrapolating, chen2021learning, zhang2021confidence, christiano2017deep, lee2021pebble}; however, we recognize that reinforcement learning may not be appropriate for physical human-robot interaction. Here the human and robot are occupying the same space, and it becomes time consuming or unsafe for the robot to test multiple trajectories through the trial-and-error process of reinforcement learning.

Recall that our intended application is manipulation tasks for robot arms. Within this setting we take advantage of the underlying robot kinematics to solve for the optimal robot trajectory. More formally, we leverage constrained optimization to convert the reward model into a robot trajectory:
\begin{equation} \label{eq:M10}
    \xi_r = \argmax_{\xi \in \Xi} \sum_{\theta \in \mathcal{E}} \sum_{s \in \xi} r_{\theta}(s) \quad \text{s.t. } \xi(0) = s_0, ~\xi(H) = s_H
\end{equation}
Here $s_0$ is the start position of the robot arm (e.g., its current position) and $s_H$ is a desired goal position. In practice, the goal position may not be known or there might not be a goal in the first place; in this case we leverage \eq{M10} without the constraint $\xi(H) = s_H$. Recent research on trajectory optimization has developed several approaches for \eq{M10} \cite{howell2019altro, schulman2014motion, gill2005snopt}. Our formalism does not rely on any specific optimizer; in our experiments we use sequential quadratic programming to solve \eq{M10} and identify the optimal robot trajectory $\xi_r$.

\p{Summarizing our Algorithm} At the start of the $i$-th interaction the robot has an ensemble of reward models with weights $\mathcal{E} = \{\theta_1, \theta_2, \ldots, \theta_m\}$. The robot applies \eq{M10} to identify the optimal trajectory under the learned rewards, and then uses shared control to track this desired trajectory $\xi_r$. The human onlooker may intervene to kinesthetically guide the robot through the task, physically correct the robot's motion, or rate the robot's overall behavior. We add this human feedback to the dataset of demonstrations $\mathcal{D}$ for the first interaction (i.e. if $i = 1$), and to the dataset corrections $\mathcal{C}$, or preferences $\mathcal{Q}$ for all other interactions ($i > 1$).The robot then updates its reward models to minimize the unified loss function in \eq{M9} --- the robot leverages these updated rewards to the start of interaction $i+1$.

\section{Simulation 1: Learning from Multiple Forms of Interaction}\label{sec:sim_interactions}
Now that we have developed a unified learning framework for physical human-robot interaction, we will compare different versions of our approach to the state-of-the-art baselines. As discussed in Section \ref{sec:related}, several approaches learn from humans using physical interactions. Some of these approaches learn end-to-end models that capture the user’s preferences for the task and learn a policy from the feedback provided, while others assume some knowledge over the features in the environment. In the latter, the features capture the task-specific concepts and are assumed to be prior knowledge of the tasks that the robot may need to perform. In this section, we perform a detailed comparison of various physical interaction approaches that learn from demonstrations, corrections and preferences, and their various combinations.\\
\p{Independent Variables} 
We test eleven algorithms that learn from physical interactions. Among these eleven algorithms, seven are different versions of Our approach, i.e. using only demonstrations (\textbf{Ours (D)}), only corrections (\textbf{Ours (C)}), only preferences (\textbf{Ours (P)}), demonstrations + corrections (\textbf{Ours (DC)}), demonstrations + preferences (\textbf{Ours (DP)}), corrections + preferences (\textbf{Ours (CP)}) and demonstrations + corrections + preferences (\textbf{Ours (DCP)}). We include three baselines that learn end-to-end networks without using any predefined features – human-gated behavior cloning (\textbf{BC}) \cite{schulman2014motion}, adversarial inverse reinforcement learning (\textbf{AIRL}) \cite{fu2017learning} and a method for learning from demonstrations and preferences developed for Atari games (\textbf{Atari}) \cite{ibarz2018reward}. We also use one baseline that assumes prior knowledge of the features in the environment and learns from a combination of demonstrations, corrections and preferences (\textbf{RRIC}) \cite{jeon2020reward}.

During \textbf{BC}, the robot learns a policy from the human’s initial demonstrations. The robot then shows the trajectory to the human and the human can intervene to physically correct and improve the robot’s behavior at any point in the trajectory. \textbf{AIRL} utilizes the demonstrations and corrections provided by the human to recover a reward function. The robot then optimizes that reward function to generate new behaviors, and compares them to the human’s original inputs. We used the repository from \cite{zhang2021confidence} to implement \textbf{AIRL}. \textbf{Atari} uses a two-step approach. First the robot leverages the human’s demonstrations to learn a policy using imitation learning methods. The robot then shows the human sample trajectories generated using the learned policy, and the human indicates their preferences. These preferences are then used to learn a reward function that we optimize by applying the soft actor-critic algorithm and generating queries in \textbf{Atari}. Finally, in \textbf{RRIC} the robot assumes full knowledge of the features in the environment and the reward function is modeled as a linear combination of these features. Based on the demonstrations, corrections and preferences provided by the human, feature weights are updated using Bayesian Inference. \textbf{Ours} directly learns a mapping from states to reward values and generates a trajectory to optimize that reward.

\begin{figure}[t]
    \centering
    \includegraphics[width=1\columnwidth]{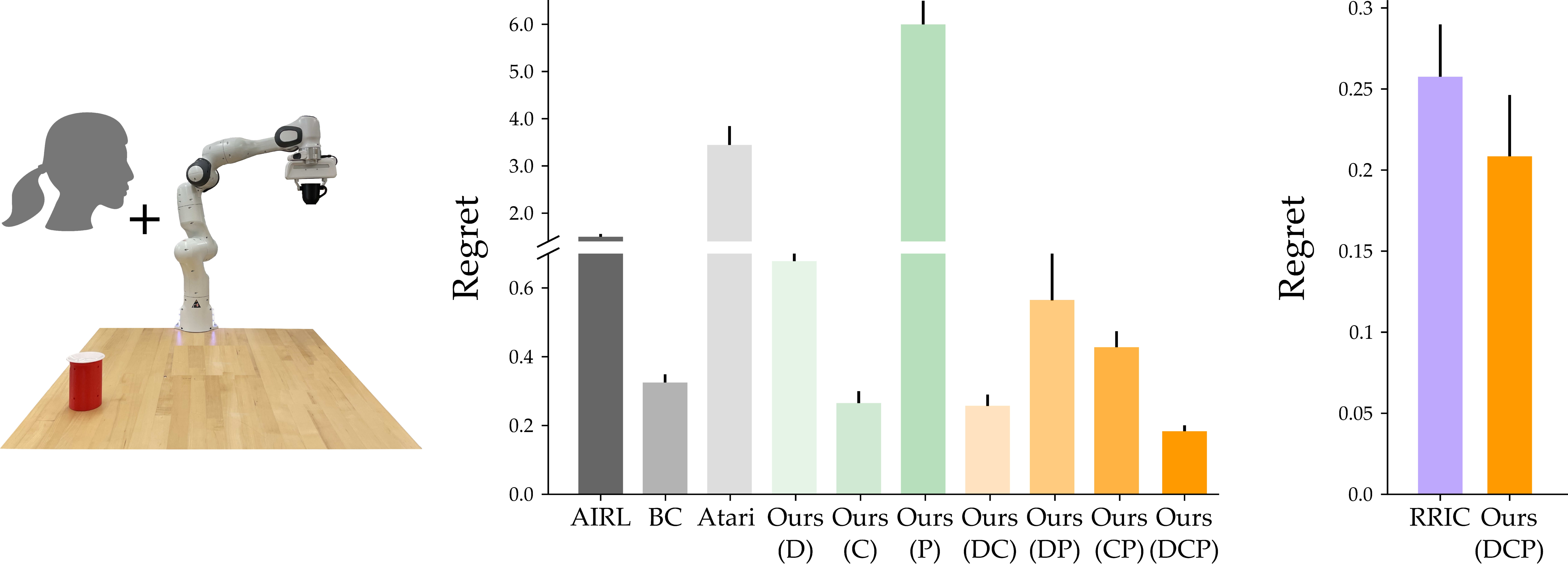}
    \caption{Experimental results for simulated humans paired with a Franka Emika robot arm. (Left) we compare different versions of our approach to state-of-the-art end-to-end learning baselines as well as a feature-based approach that combines multiple forms of feedback. (Center) $15$ simulated humans perform each task (\textit{Laptop} and \textit{Table}) using all the end-to-end learning algorithms. (Right) The simulated humans perform each task with a feature-based learning algorithm. We record the performance of the robot after learning from each approach in the form of regret and report the average regret and standard error. \textbf{Ours (DCP)} significantly outperforms all other versions of our approach ($p<.05)$. \textbf{Ours (DCP)} has a significantly lower average regret as compared to the end-to-end learning methods ($p<.05$) and performs at par with \textbf{RRIC}. We emphasize that \textbf{RRIC} has access to all relevant features in the environment, while \textbf{Ours} learns the reward function from scratch.}
    \label{fig:sim1_new}
\end{figure}

\p{Procedure}
The simulated human and a simulated robot performed two tasks (\textit{Table} and \textit{Laptop}) with each algorithm (see Fig. \ref{fig:sim1_new}). For each of these tasks, the simulated human is teaching the robot to carry a cup of coffee to a goal position. In \textit{Table}, the human wants the robot to carry the cup of coffee close to the table; in \textit{Laptop}, the human wants the robot to avoid going over the laptop while moving to the goal location. For this experiment we used a 7-DoF Franka-Emika Panda robot arm. The robot \textit{did not have access} to the task reward function. The simulated human knew their reward function and provided demonstrations, corrections, and preferences to optimize that reward and teach the robot.

Within this experiment we kept the number of interactions between the simulated human and the robot constant for each method. For \textbf{BC} and \textbf{AIRL}, the simulated human provided 6 demonstrations. For \textbf{Atari} the human first provided 2 demonstrations and then asked for 4 preferences to the human. \textbf{RRIC} received an even split for each feedback type: 2 demonstrations, 2 corrections and 2 preferences. Similarly, all different versions of \textbf{Ours} received an even split for each feedback type that version is meant to incorporate. For example, \textbf{Ours (D)} was given 6 demonstrations and \textbf{Ours (DC)} used 3 demonstrations and 3 corrections.
\\
\p{Dependent Variables}
The simulated human worked with the robot to provide feedback across 6 interactions, and the robot optimized its learned reward to produce its final trajectory. We measured the performance of the robot by computing the \textit{regret} of this learned trajectory:
\begin{equation}\label{eq:s1}
    Regret(\xi) = \sum_{s\in \xi^*}r_{\theta^*}(s) - \sum_{s\in \xi}r_{\theta^*}(s)
\end{equation}
where $r_{\theta^*}$ is the true reward function for the task, $\xi^*$ is the optimal robot trajectory for the task, and $\xi$ is the robot's learned trajectory. Regret quantifies how much worse the robot's learned behavior is than the ideal behavior: lower values of regret indicate better robot performance.\\

\p{Hypothesis}
For this simulation, we had the following hypotheses:\\
\textbf{H1}: \textit{Our proposed approach with demonstrations, corrections, and preferences, \textbf{Ours (DCP)}, will outperform our approach with only one or two types of feedback.}\\
\textbf{H2}: \textit{\textbf{Ours (DCP)} will outperform the end-to-end learning baselines.}\\
\textbf{H3}:\textit{ The performance of \textbf{Ours (DCP)} will match \textbf{RRIC} with known features.}\\
\p{Results}
Our results for this simulation are summarized in Fig. \ref{fig:sim1_new}. Performing a repeated measures ANOVA test (Normality of data verified using Q-Q plots) with a Greenhouse-Geisser correction, we found that the robot’s learning algorithm had a significant effect on the regret ($F(1.818, 140) = 61.939$, $p < 0.05$). By contrast, we found that the task did not have a significant effect on the regret ($F(1, 14) = 3.073$, $p = 0.101$). Thus, we report the combined regret for both the tasks in our results.

From the regret plots in Fig. \ref{fig:sim1_new} (Center), we observe that by combining all three forms of feedback our approach outperforms all other versions of \textbf{Ours} where we use only one or two forms of feedback ($p < 0.05$). This provides support for our hypothesis \textbf{H1}. We also observe that \textbf{Ours (C)} and \textbf{Ours (DC)} have a lower regret compared to \textbf{Ours (D)}, \textbf{Ours (P)}, \textbf{Ours (DP)} and \textbf{Ours(CP)}. This is a result of the iterative nature of the corrections as compared to the demonstrations, where the human provides all inputs at once before the robot updates its reward model.

We also notice that the end-to-end learning approaches perform at par with or better than some versions of \textbf{Ours} when only one or two types of feedback are available to our approach. While \textbf{Atari} and \textbf{Ours (DP)} receive the same forms of feedback, the regret for \textbf{Atari} is higher owing to the limited amount of data available for training. We observe that \textbf{BC} has a lower regret than \textbf{Ours (D)}, but performs at par with \textbf{Ours (C)}. This suggests that the performance of \textbf{Ours} improves when it receives incremental feedback from humans. However, when all three forms of feedback are made available to our approach, \textbf{Ours (DCP)} significantly outperforms all the end-to-end learning models ($p < 0.05$). This suggests that learning from multiple types of feedback is more effective than learning from just one or two types of feedback. Here, we find support for \textbf{H2}.

Finally, we compare \textbf{Ours (DCP)} to \textbf{RRIC}, which has access to the features in the environment
and learns using all three forms of feedback within a Bayesian inference framework (see Fig. \ref{fig:sim1_new} (Right)). We observe that the performance of \textbf{Ours (DCP)} is comparable to \textbf{RRIC} ($p = 0.548$). This suggests that \textbf{Ours} — an approach that learns the reward end-to-end without any features — can perform as well as a Bayesian Inference approach that requires prior knowledge of all the features. Here we find support for our hypothesis \textbf{H3}.

\section{User Study: Multiple Forms of Physical Interaction} \label{sec:user}

So far we have evaluated our method in controlled experiments with simulated human users. In this section we will test our unified approach on \textit{actual} participants. These participants physically interact with a $7$-DoF robot arm by applying forces and torques, and communicate their intended task to the robot through demonstrations, corrections, and preferences. We compare our algorithm to interactive learning baselines that combine multiple types of human feedback. Here we explore scenarios where the robot must learn the task from scratch: our method and the baselines have no prior knowledge of the features or the tasks that the participants want to complete.  Users must communicate their desired tasks through physical interaction. Videos of our user study are available at \href{https://youtu.be/FSUJsTYvEKU}{https://youtu.be/FSUJsTYvEKU}.

\begin{figure}
    \centering
    \includegraphics[width=1\columnwidth]{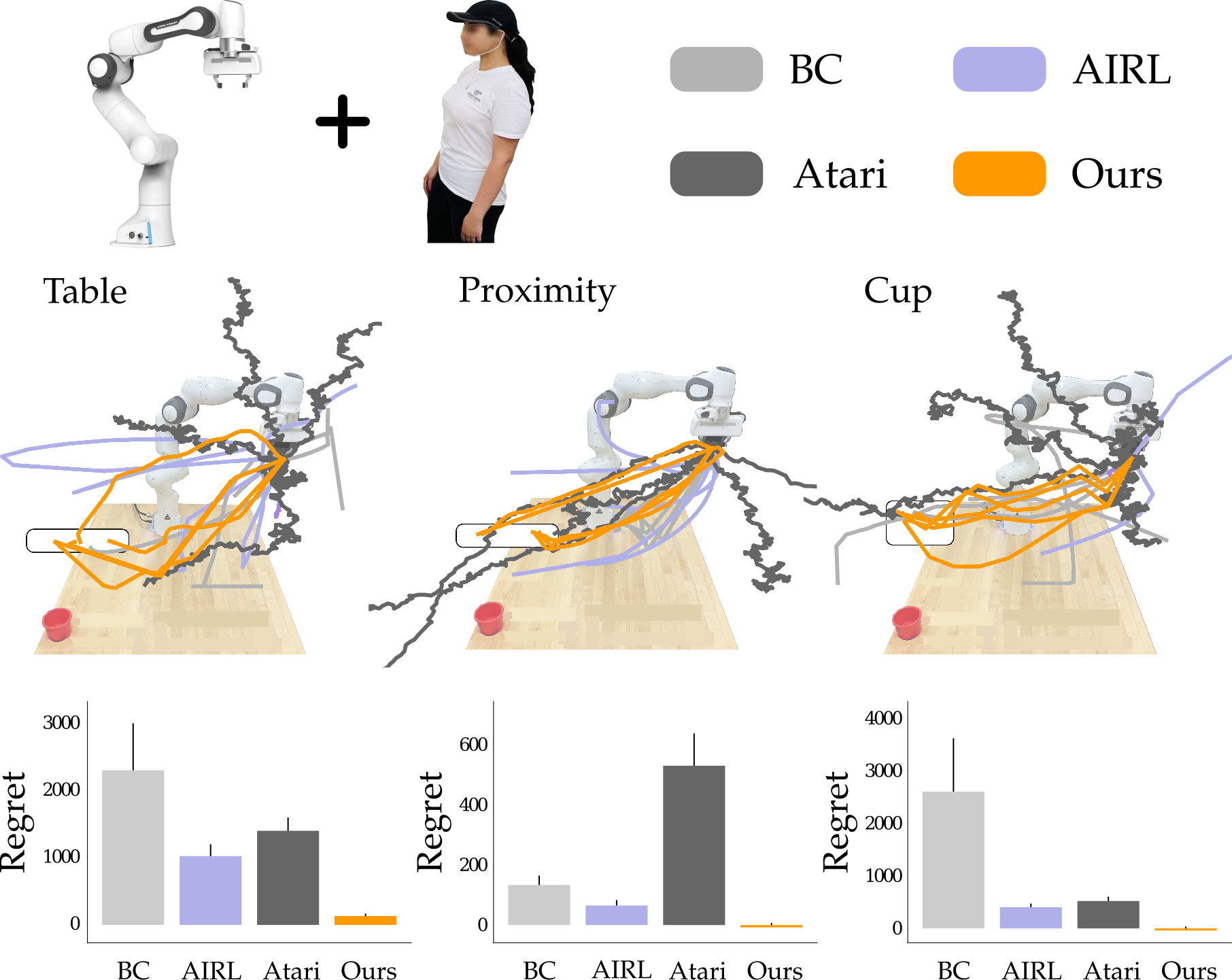}
    \caption{Learned trajectories and objective results from our in-person user study. (Top) Participants physically interacted with a $7$-DoF robot arm that had no prior knowledge about the tasks. The robot learned from physical interactions using our approach and imitation learning baselines that combine multiple feedback modalities. (Middle) The final trajectories the robot learned with each method. Five users taught the robot the \textit{Table} task, five users taught the \textit{Proximity} task, and five users taught the \textit{Cup} task. During each task the robot needed to reach a goal position within the white rectangle. We trace the $xyz$ position of the robot's end-effector; within the \textit{Cup} task the robot also needed to maintain specific orientations. (Bottom) The regret between the robot's learned trajectory and ideal trajectory. Lower values of regret indicate that the robot completed the task correctly, and the error bars plot standard error of the mean. \textbf{Ours} outperforms \textbf{AIRL} and \textbf{Atari} on the \textit{Table} and \textit{Cup} tasks ($p<.05$), and \textbf{Ours} has a lower regret than all the baselines for the \textit{Proximity} task ($p<.05$).}
    \label{fig:user_study}
\end{figure}

\p{Independent Variables} We tested four different algorithms that learn from physical interaction. Similar to Section \ref{sec:sim_interactions}, our baselines include human-gated behavior cloning (\textbf{BC}) \cite{ross2011reduction}, adversarial inverse reinforcement learning (\textbf{AIRL}) \cite{fu2017learning}, and a method for learning from preferences and demonstrations developed for Atari games (\textbf{Atari}) \cite{ibarz2018reward}. \textbf{Ours} leverages the unified algorithm introduced in Section~\ref{sec:method}. We emphasize that each of these approaches learns without pre-defined features. The implementation of each of these approaches follows the procedure described in Section \ref{sec:sim_interactions}.

\p{Experimental Setup} Participants physically interacted with a $7$-DoF Franka Emika robot arm. During the user study humans tried to teach this robot three tasks (see \fig{user_study}). In \textit{Table} the robot had to reach the goal while carrying a cup close to the table; in \textit{Proximity} the robot had to move to a goal region while staying away from the user. Note that the nature of these two tasks is similar to that of the experiments performed in Section \ref{sec:sim_interactions}. In this user study we introduce a new task, \textit{Cup}, where the participants teach the robot to complete a scooping action and then pour the cup at the goal position. We asked participants to mark their goal at the start of each interaction. To encourage more diverse human inputs, participants were instructed to change their goal position within a marked region between interactions.
 
\p{Participants and Procedure} For our user study we recruited $15$ participants from the campus community ($5$ female, average age $25 \pm 4$ years). Participants gave informed written consent prior to the start of the experiment under Virginia Tech IRB $\#22$-$308$.

The participants were divided into three groups of five people. Each group of participants performed a single task (i.e., participants only taught the table task, the proximity task, or the cup task). Importantly, users taught this task with \textit{all four} of the robot learning algorithms. The order of the algorithms was counterbalanced using a Latin square design: e.g., some participants began with \textbf{Ours}, and others began with \textbf{BC}. For each learning algorithm the human and robot started over from scratch: the robot had no prior information, and the human provided new demonstrations, corrections, or preferences to convey their task.

For \textbf{BC} and \textbf{AIRL} the human provided $6$ demonstrations\footnote{For \textbf{BC} and \textbf{AIRL} we gave users the option of providing corrections that only modify a segment of the robot's behavior. However, since the robot's learned behavior after two demonstrations was far from the user's intended task, participants chose to keep demonstrating the entire trajectory. Note that neither \textbf{BC} or \textbf{AIRL} can learn from preferences.}. With \textbf{Atari} the participant first provided $2$ demonstrations and then the robot asked for $4$ preferences (to reach a total $6$ interactions). Finally, observing the result from Section \ref{sec:sim_interactions}, that multiple forms of feedback enable a better learning, we provide our approach with all three forms of feedback. We divide \textbf{Ours} evenly between each type of physical feedback: users gave two demonstrations, corrections, and preferences (to maintain a total of $2+2+2=6$ interactions).

\p{Dependent Variables} After participants finished providing their inputs, the robot leveraged its learning algorithm to identify a final trajectory. We measured how effectively this final trajectory completed the intended task. More specifically, we applied \ref{eq:s1} to quantify the regret between the robot's actual behavior and the ideal task behavior.

We also administered a 7-point Likert scale survey \cite{schrum2020four} to assess the participants' subjective responses to each learning condition. Our survey questions were organized into four multi-item scales: how \textit{easy} it was to physically provide feedback to the robot, whether the robot \textit{learned} to perform the task correctly, how \textit{flexible} the robot was to different types of physical interaction, and if they would \textit{prefer} using this method in the future. Every participant completed this survey four times: once after they finished working with each robot learning algorithm.

\p{Hypothesis} For our user study we had the following hypotheses:\\
\textbf{H4}: \textit{Robots using our unified learning approach will perform the task better after the same number of physical human interactions.}\\
\textbf{H5}: \textit{Participants will subjectively prefer our learning algorithm as compared to the baselines.}

\floatname{table}{\color{blue} Table}

\begin{table}[]
\caption{Questions on our Likert scale survey. We grouped questions into four scales and tested their reliability using Cronbach’s $\alpha$. We explored whether providing feedback to the robot was easy, the robot learned the task, if the users liked the flexibility of using different feedback forms, and if they preferred to use the method in future. Computed $p$-values indicate if users preferred our approach to the baselines, where $*$ denotes statistical significance.}
\label{tab:likert}
\resizebox{\columnwidth}{!}{%
\begin{tabular}{@{}lccccc@{}}
\toprule
\multirow{2}{*}{Questionnaire Item} &
  \multirow{2}{*}{Reliability} &
  \multirow{2}{*}{F(3,42)} &
  \multicolumn{3}{c}{p-value} \\ \cmidrule(l){4-6} 
 &
   &
   &
  BC &
  AIRL &
  Atari \\ \midrule
—It was easy to provide feedback to the robot. &
  \multirow{2}{*}{0.863} &
  \multirow{2}{*}{1.707} &
  \multirow{2}{*}{0.486} &
  \multirow{2}{*}{0.7} &
  \multirow{2}{*}{0.185} \\
—Providing feedback to the robot was challenging. &
   &
   &
   &
   &
   \\ \midrule
— The robot learned to perform the task correctly. &
  \multirow{2}{*}{0.911} &
  \multirow{2}{*}{11.982} &
  \multirow{2}{*}{p \textless 0.05*} &
  \multirow{2}{*}{p \textless 0.05*} &
  \multirow{2}{*}{p \textless 0.05*} \\
\begin{tabular}[c]{@{}l@{}}— The robot's motion did not align with what \\ I was trying to teach the robot.\end{tabular} &
   &
   &
   &
   &
   \\ \midrule
—I liked showing different types of feedback. &
  \multirow{2}{*}{0.789} &
  \multirow{2}{*}{0.225} &
  \multirow{2}{*}{0.689} &
  \multirow{2}{*}{0.571} &
  \multirow{2}{*}{0.427} \\
—I preferred just showing one type of feedback repeatedly. &
   &
   &
   &
   &
   \\ \midrule
— I would use this method in the future. &
  \multirow{2}{*}{0.806} &
  \multirow{2}{*}{4.263} &
  \multirow{2}{*}{0.353} &
  \multirow{2}{*}{p \textless 0.05*} &
  \multirow{2}{*}{p \textless 0.05*} \\
\begin{tabular}[c]{@{}l@{}}— I would prefer another approach that I tried \\ if I was to do this again.\end{tabular} &
   &
   &
   &
   &
   \\ \bottomrule
\end{tabular}%
}
\end{table}

\begin{figure}
    \centering
    \includegraphics[width=0.75\columnwidth]{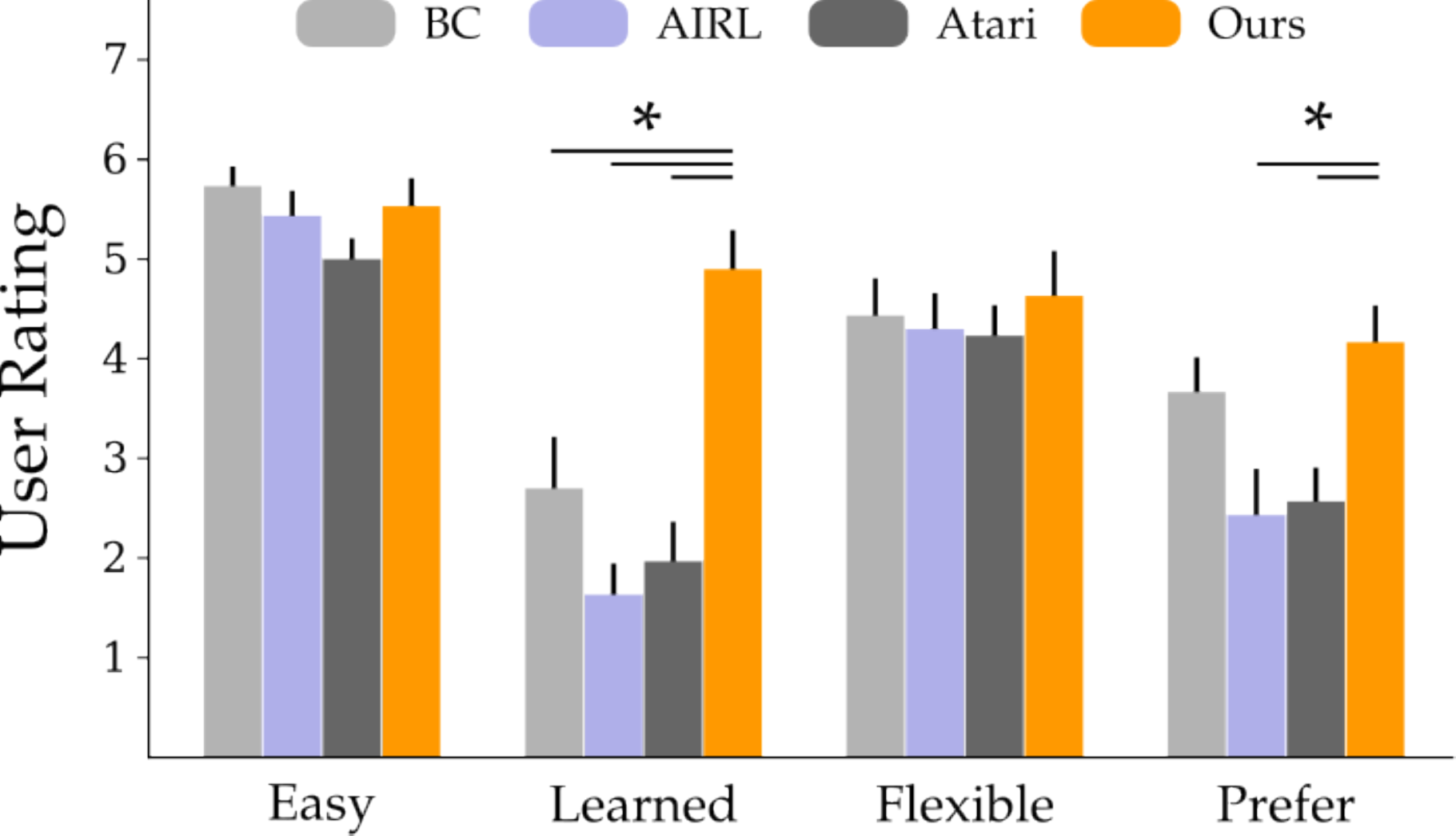}
    \caption{Subjective results from our in-person user study. Higher ratings indicate user agreement (e.g., a score of $7$ indicates that it was \textit{easier} to provide physical feedback). Error bars show standard error of the mean, and an $*$ denotes statistical significance ($p<.05$). After watching the final trajectory learned by each approach, participants rated \textbf{Ours} as a better \textit{learner} than the baselines. Users also \textit{preferred} \textbf{Ours} to \textbf{AIRL} and \textbf{Atari}.}
    \label{fig:likert}
\end{figure}

\p{Results} The objective results from our user study are presented in \fig{user_study}, and the subjective responses are summarized in \fig{likert}. Let us start with the objective results: After verifying the normality of the data using Q-Q plots and performing a repeated measures ANOVA test on our results, we found that the robot's learning algorithm had a significant main effect on regret ($F(3,12)=6.942$, $p<.05$). Looking at \fig{user_study} we notice that the regret for \textbf{Ours} is consistently lower than the baselines. Here lower regret is better --- this indicates that the robot learned trajectories better matched the human's intended task. We can directly observe this trend from the final trajectories shown above: notice that \textbf{Ours} consistently moves to the goal region and completes the task, while the alternatives produce noisy, inconsistent motions. Post hoc comparisons confirm that \textbf{Ours} outperformed \textbf{AIRL} and \textbf{Atari} on the \textit{Table} and \textit{Cup} tasks ($p<.05$), and that \textbf{Ours} had an overall lower regret than all baselines for \textit{Proximity} task ($p < .05$). We observe that given the limited amount of data (only 6 interactions), \textbf{Ours} has lower average regret with a low variance across all three tasks. On the other hand, \textbf{Atari} and \textbf{BC} fail to learn the task representations accurately from the same limited amount of data, and thus have a higher variance in their performance. Overall, the results shown here and in the video submission indicate that our unified learning approach was best able to synthesize the human's physical inputs and learn the correct task from scratch.

Next we consider the results from our Likert scale survey in \fig{likert} and Table \ref{tab:likert}. After confirming that the scales were reliable (Cronbach's $\alpha > 0.7$), we grouped each scale into a single combined score and performed a one-way repeated measures ANOVA on the results. When users watched the robot's final learned behavior they perceived \textbf{Our} approach as a better learner than the baselines ($F(3, 42) = 11.982$, $p<.05$), and when they considered their experience teaching the robot they preferred to use \textbf{Ours} over the \textbf{AIRL} and \textbf{Atari} approaches ($F(3,42) = 4.263$, $p<.05$). One confounding factor here is the about of time it took for the robot to learn from human's demonstrations. With \textbf{Ours} and \textbf{BC} the entire process from teaching the robot to autonomously completing the task took roughly $10$ minutes, while with \textbf{AIRL} and \textbf{Atari} it took more than $15$ minutes on average (this additional time was needed to train the robot's policy). Participants may have preferred \textbf{Ours} and \textbf{BC} in part because they completed the training process more quickly. However, our overall results support hypothesis \textbf{H5} and suggest that participants subjectively perceived our unified approach as a better learner from demonstrations, corrections, and preferences.

\section{{Simulation 2}: Learning with Known and Unknown Features} \label{sec:sim}

Now that we have tested our approach against baselines that learn end-to-end models, we will compare our approach to state-of-the-art baselines that use the knowledge about the features in the environment to learn the reward weights. As we discussed in Section~\ref{subsec:prelim}, several related works learn from combinations of demonstrations, corrections, and preferences by assuming that the reward function is based on features. These features capture task-relevant concepts (e.g., the orientation of the chair leg) and are programmed using prior knowledge of the tasks the robot will need to perform. Here we consider situations in which the robot must learn \textit{expected} tasks --- i.e., cases where the features apply --- as well as \textit{unexpected} tasks where the pre-programmed features are insufficient. We conduct these experiments with simulated humans that provide inputs to real robot arms. Overall, we break this section down into two parts: (a) a comparison to physical interaction approaches that learn from demonstrations and corrections, and (b) a comparison to learning approaches that combine demonstrations and preferences.

\subsection{Learning from Demonstrations and Corrections} \label{sec:sim1}
\p{Independent Variables} We consider two baselines for learning from physical demonstrations and corrections: \textbf{Coactive} \cite{losey2021physical, jain2015learning} and \textbf{FERL} \cite{bobu2022inducing}. In \textbf{Coactive} the robot assumes that it knows \textit{all} the features for the current task; based on the human's inputs, the robot builds a reward function from these features and then selects the optimal trajectory. By contrast, in \textbf{FERL} the robot recognizes that it may be missing some task-related features. Here the robot leverages the feature demonstrations (feature traces) provided by the human to first learn the unknown features --- once it has a model of these features, it then applies the same method as \textbf{Coactive} to build the reward function. Remember that our proposed approach (\textbf{Ours}) does not rely on features. Instead, we learn a mapping directly from states to rewards; for cases where the features are given, \textbf{Ours} can incorporate those features in the augmented state to be used as an input to our reward model (for learning and execution). Note that here, our reward model has the same information about the features as the other approaches (i.e. if \textbf{Coactive} and \textbf{FERL} have information about only \textit{1 feature}, \textbf{Ours} also has information about only one feature).

\begin{figure}
    \centering
    \includegraphics[width=1\columnwidth]{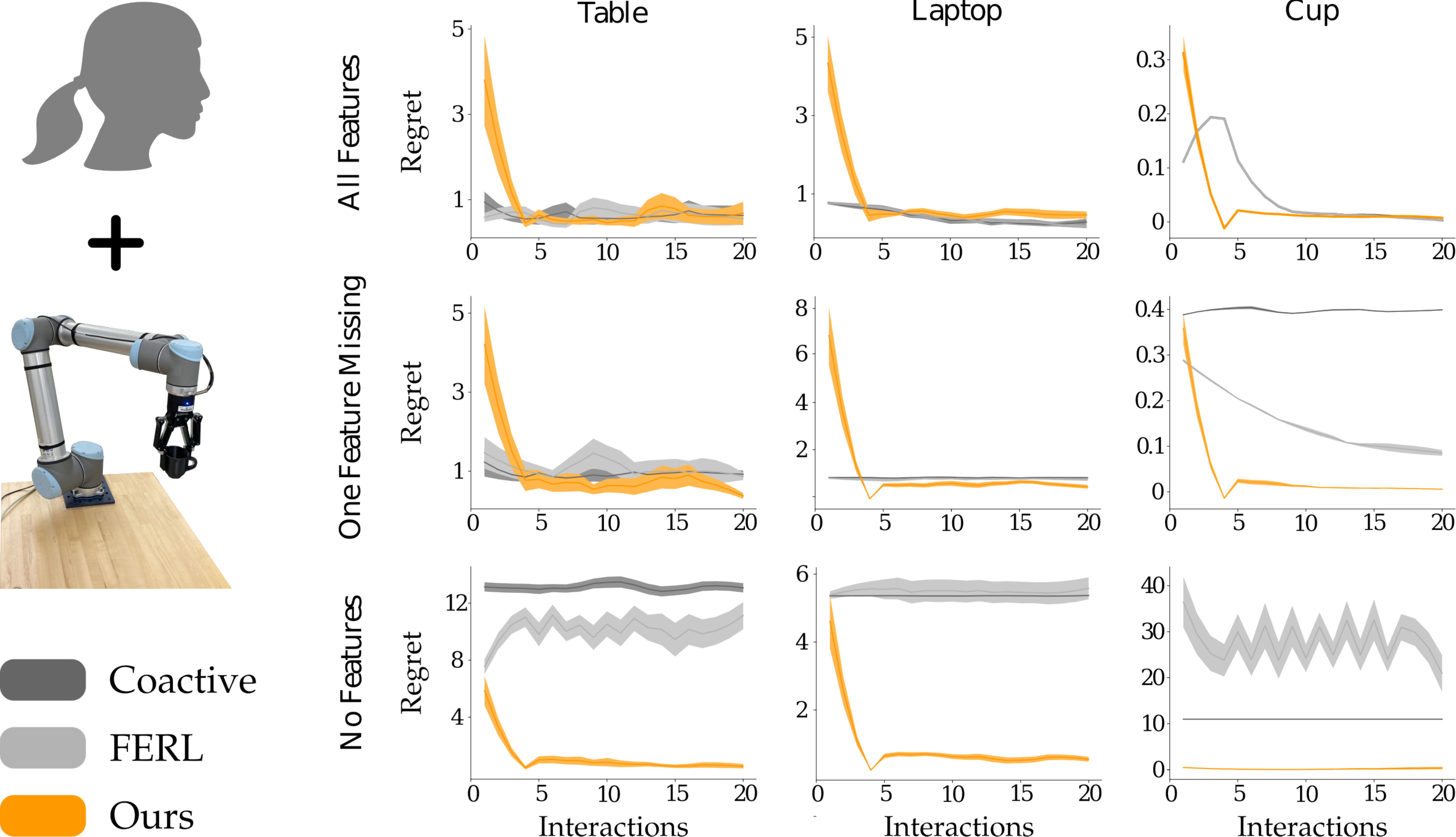}
    \caption{Experimental results for simulated humans paired with a UR10 robot arm. (Left) We compare our approach to two existing algorithms that learn from physical interaction. \textbf{Coactive} \cite{losey2021physical, jain2015learning} assumes that the reward is composed of pre-programmed features, and \textbf{FERL} \cite{bobu2022inducing} learns features from human demonstrations before constructing a reward function from those features. (Right) Over repeated interactions $10$ simulated humans input demonstrations and corrections to teach the \textit{Table}, \textit{Laptop}, and \textit{Cup} tasks. The columns correspond to the tasks, and the rows capture the prior information the robot has about each task. In the first row the robot is given all task-related features, in the middle row the robot is missing one feature, and in the bottom row the robot has no prior information about the task. The plots show regret (the difference in reward between the ideal trajectory and the learned trajectory), and the shaded regions show the standard error. At the end of all $20$ interactions \textbf{Ours} performs similar to or worse than \textbf{Coactive} and \textbf{FERL} when all features are known. If one or more feature is missing, however, \textbf{Ours} outperforms both baselines.} 
    \label{fig:sim1}
\end{figure}

\p{Procedure} The simulated human and real robot performed three tasks with each learning algorithm (see \fig{sim1}). For all three tasks the human is teaching the robot to reach a goal position. In \textit{Table} the human wants the robot arm to move close to the table; in \textit{Laptop} the human wants the robot arm to avoid passing above a laptop; and in \textit{Cup} the human wants the robot arm to carry a cup upright so that it does not spill. Each task has two potential features: the goal that the robot should reach (e.g., distance to the goal) and the way the robot arm should move towards that goal (e.g., height from the table, distance from the laptop, or orientation of the cup). For these experiments we used a $6$-DoF UR10 robot arm. This robot \textit{did not know} the task reward function. To teach the real robot in a controlled setting, we used a simulated human: the simulated human knew the correct reward, and provided demonstrations or corrections that optimized this reward.

We seek to understand how our approach compares to baselines both when the robot has prior knowledge about the task and when the task is new or unexpected. Accordingly, we tested three different conditions: (a) when the robot knows all task-related features, (b) when one task-related feature --- \textit{Laptop}, \textit{Table} or \textit{Cup} --- is missing, and (c) when the robot does not know any features of the task.

\p{Dependent Variables}  The simulated human worked with the real robot to provide 20 inputs (i.e., demonstrations and corrections) over repeated interactions. For \textbf{Coactive} and \textbf{Ours} the first input is a task demonstration, and the remaining interactions are corrections. For \textbf{FERL} the type of input depends on the number of unknown features. When all the features are given, \textbf{FERL} also starts with a task demonstration followed by corrections. But when any feature is missing, the human first provides 10 feature demonstrations per missing feature. After these offline demonstrations (which are meant to teach features to the robot), the human provides one task demonstration and corrections similar to \textbf{Coactive} and \textbf{Ours}. Thus, each method receives one task demonstration and 19 corrections to learn the task (the feature demonstrations are not included as part of the 20 interactions). After each interaction we measured the performance of the learning robot using equation \ref{eq:s1}.

\p{Hypothesis} For this experiment we had the following hypotheses:\\
\textbf{H6}: \textit{Our method will match the baselines when all the features are known.}\\
\textbf{H7}: \textit{Our method will outperform both \textbf{Coactive} and \textbf{FERL} when one or more features are missing.}

\p{Results} Our results are visualized in \fig{sim1}. Note that this figure is a grid: the columns are the tasks, and the rows are the amount of prior knowledge that the robot has about the tasks. To analyze the results we first verified the normality of our data using Q-Q plots and then performed a repeated measures ANOVA with Greenhouse-Geisser correction. We found that the robot's learning algorithm had an effect on regret across all tasks and conditions: $F(1.040, 18)=29.063$, $p<0.05$.

To understand this result we next explored how the robot's performance changed based on the amount of prior information available to the learning algorithm. In the first row of \fig{sim1} we consider scenarios where all the task-related features are known. For example, during the \textit{Table} task the robot knows that the reward is a function of the distance from the goal and the height of the end-effector. Overall, we found that --- when all features are given --- \textbf{Ours} performs on par with or worse than the baselines by the end of all $20$ interactions. Post hoc tests revealed that for the \textit{Table} task \textbf{Ours} matched the alternatives (\textbf{Coactive}: $p=0.788$, \textbf{FERL}: $p=0.790$). During \textit{Laptop} our method performed similarly to \textbf{Coactive} ($p=0.117$) but had higher regret than \textbf{FERL} ($p<0.05$). Finally, within \textit{Cup} both \textbf{FERL} and \textbf{Coactive} outperformed our approach at the last interaction ($p<0.05$) These results partially support hypothesis \textbf{H6}: when the robot is given prior information about all the features, existing methods leverage this structure to accurately learn the human's reward. Our approach starts with worse performance because it does not make assumptions about the reward function and must learn to focus on the given features.

However, the relative performance changes once the robot encounters new or unexpected tasks. In the second row of \fig{sim1} we test settings where the robot knows one feature (distance to the goal) but does not know the other task-related feature. Because \textbf{Coactive} assumes that all the features are given, it treats each human input as an observation about the correct distance to the goal (and never realizes that the human's inputs may be communicating something else). \textbf{FERL} takes a step towards resolving this problem by trying to learn the missing feature from the first four interactions. But post hoc tests reveal that our proposed learning approach matches or outperforms the feature-based alternatives.  For the \textit{Table} task there are no statistically significant differences between \textbf{Ours} and\textbf{Coactive} ($p=0.074$), but \textbf{Ours}  has a lower regret than \textbf{FERL} ($p<0.05$). On the other hand, during \textit{Laptop} and \textit{Cup} our method leads to less regret than both the baselines by the end of the physical interactions($p<0.05$). This trend continues in the final row of \fig{sim1} where the robot has no prior information about the task reward. Here \textbf{Ours} outperforms both baselines across all three tasks ($p<0.05$). Overall, these results suggest that when the robot encounters situations where it has incomplete information, our unstructured reward learning approach is better able to capture the correct task than baselines that depend on task-related features. We therefore find support for \textbf{H7} when the robot is learning from physical demonstrations and corrections.

\subsection{Learning from Demonstrations and Preferences} \label{sec:sim2}

\p{Independent Variables} So far we have compared our approach to baselines developed specifically for physical interaction when learning from demonstrations and corrections. Next, we turn our attention to alternate approaches that learn from demonstrations and preferences \cite{ibarz2018reward, brown2019extrapolating, chen2021learning, zhang2021confidence, jeon2020reward}: although these methods are not designed only for physical interaction, they can be applied to our setting. Here we compare \textbf{Ours} to \textbf{DemPref} \cite{biyik2021learning}, a recent approach that combines both demonstrations and preferences to build a model of the human's reward function. Like \textbf{Coactive} in the previous experiment, \textbf{DemPref} assumes that the reward function is composed of features, and the robot knows all the relevant features for the current task. In this experiment, we aim to study the trade-off between our proposed approach and approaches that utilize Bayesian inference to learn the task representation from human feedback.

The \textbf{DemPref} algorithm has two parts. First, the robot gets demonstrations from the human to learn a rough estimate of the reward function; next, the robot actively queries the human to elicit their preferences and fine-tune the learned reward. Recall from Section~\ref{sec:method} that under our proposed approach the robot can collect human preferences passively or actively. We will therefore consider two different versions of \textbf{Ours}: one where the robot gets the human's preferences from randomly chosen trajectories, \textbf{Ours (Passive)}, and one where the robot applies \eq{M7} and \eq{M8} to select preference queries that will maximize the information the robot gains about $\theta$, \textbf{Ours (Active)}. To make the comparison as fair as possible, we have given both \textbf{DemPref} and \textbf{Ours} the same dataset of $1000$ trajectories from which to choose their preference queries. Each query in this dataset was sampled by choosing a random goal in the robot's workspace followed by generating two noisy trajectories to the goal. Finally, to show that obtaining human preferences improves the performance of our approach, we also include \textbf{Ours (Demo)}, a baselines where the robot learns from only one demonstration (without ever considering the human's preferences).

\begin{figure}
    \centering
    \includegraphics[width=1\columnwidth]{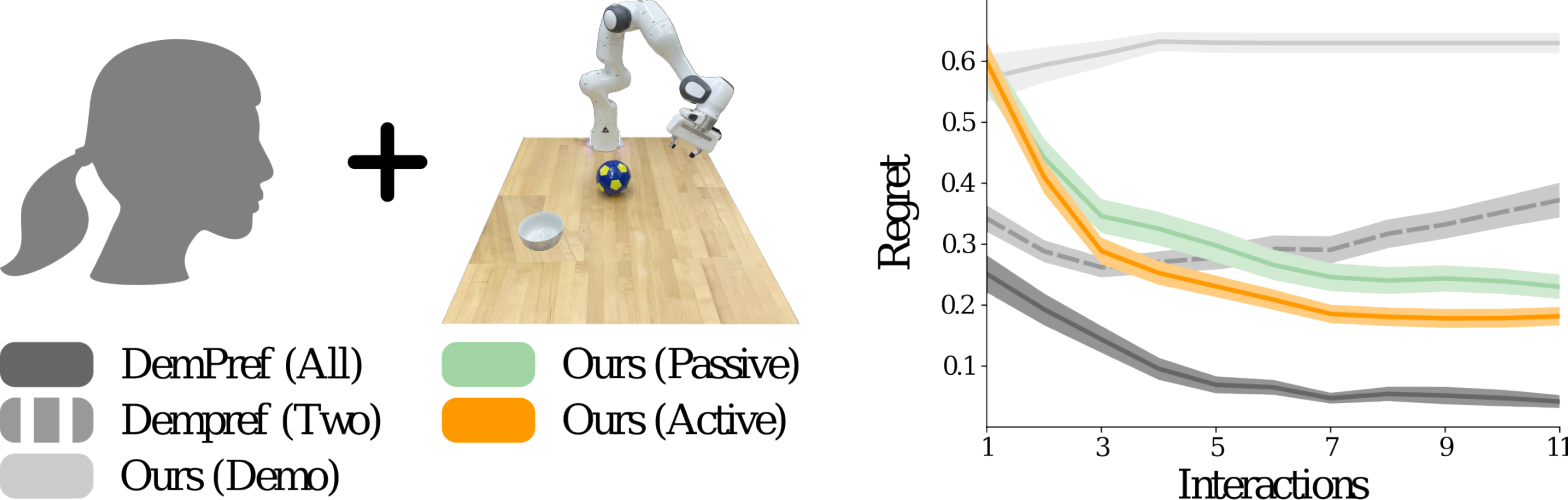}
    \caption{Experimental results for simulated humans paired with a Franka Emika robot arm. (Left) We compare our approach to \textbf{DemPref} \cite{biyik2021learning}, a method for learning from demonstrations and preferences that assumes the robot has access to all task-related features. (Right) 500 simulated humans attempt to teach the robot their desired task: each user provides one demonstration followed by $10$ preferences. In \textbf{DemPref (All)} the robot knows all three task-related features, in \textbf{DemPref (Two)} the robot is missing one feature, and in \textbf{Ours (Demo)} the robot only observes a single demonstration. We compare these baselines to our approach when the robot chooses preference queries at random, \textbf{Ours (Passive)}, and when the robot asks questions to gain as much information as possible, \textbf{Ours (Active)}. The shaded region is the standard error. \textbf{Our} approach outperforms \textbf{DemPref} when a feature is missing and the robot must learn an unexpected task.}
    \label{fig:sim2}
\end{figure}

\p{Procedure} We perform this experiment in a controlled environment by pairing simulated humans with a $7$-DoF Franka Emika robot arm (see \fig{sim2}). The environment has three features: the distance of the robot from the bowl, the height of the robot from the table, and the distance between the robot and the ball. For each learning algorithm we simulated $500$ humans with randomly selected reward functions that depend on these three features. Put another way, every simulated human assigns a different relative importance to the task features following \eq{P1}. The robot does not know the human's reward function \textit{a priori}. Over repeated interactions, the robot attempts to identify the correct reward (and the corresponding optimal trajectory) from the demonstrations and preferences of the current human user. Each simulated user selects their inputs to noisily optimize their internal reward function.

This experiment is designed similarly to the simulation in Section~\ref{sec:sim1}. We want to explore how our approach compares to \textbf{DemPref} in situations where the robot encounters a familiar task and settings where the robot is faced with new or unexpected tasks. We therefore compare \textbf{Ours} to \textbf{DemPref} (a) when all the features are known and (b) when one feature is missing. Recall that the task has three potential features. For case where one feature is missing, we performed separate trials where we removed either the first feature, the second feature, or the third feature; we then report the average across these runs. \textbf{Our} approach was never given any information about the features (i.e., \textbf{Ours} had no prior information about the task).

\p{Dependent Variables} The simulated human worked with the real robot over $11$ interactions. During the first interaction the human provides a demonstration, and during the next $10$ interactions the robot collects preferences. After each interaction the robot solves for its best guess of the task trajectory: we measure the performance of the robot learner using \textit{regret} as defined in \ref{eq:s1}.

\p{Hypothesis} For this experiment we had the following hypotheses:\\
\textbf{H8}: \textit{Our method will outperform \textbf{DemPref} when the robot does not know all the task-related features.}\\
\textbf{H9}: \textit{Our method will converge to the correct trajectory more rapidly when choosing active preference queries as compared to passively collecting preferences or ignoring preferences altogether.}

\p{Results} We summarize the results from this simulation in \fig{sim2}. Remember that lower regret scores are better: After testing for the normality of data using Q-Q plots, a repeated measures ANOVA with Greenhouse-Geisser correction revealed that the learning algorithm had a significant main effect on regret by the end of the interactions ($F(1.994, 1996)=366.897, p<.05$).

In settings where the robot encounters an expected task --- i.e., when all the features of the environment are pre-programmed into the robot arm --- \textbf{DemPref (All)} outperforms our proposed approach; \textbf{Ours (Passive)}: $p<.05$ and \textbf{Ours (Active)}: $p<.05$. However, when the robot is missing a task-related feature \textbf{DemPref (Two)} becomes confused by the human's feedback. Looking at \fig{sim2}, we notice that \textbf{DemPref (Two)}'s performance decreases as the the human provides additional preferences: this occurs because the robot is misinterpreting the human's inputs as feedback about the two known features (instead of the one missing feature). By the end of the physical interactions our proposed approach better understands the unexpected task than the baseline; \textbf{Ours (Passive)}: $p<.05$ and \textbf{Ours (Active)}: $p<.05$. Overall, the results here agree with hypothesis \textbf{H8}. From this result, we conclude that when the robot has access to the environment features, Bayesian inference approaches perform on par with or better than our proposed reward learning approach. However, in more open-ended cases where information on reward functions is missing, \textbf{Ours} is more suitable for robot learning.

We next compared different versions of our learning algorithm. First, we find that learning from both demonstrations and preferences provides more information about the task than learning only from the human's initial demonstrations. Post hoc tests show that \textbf{Ours (Demo)} has significantly higher regret than \textbf{Ours (Passive)} ($p<.05$) and \textbf{Ours (Active)} ($p<.05$). Next, we tested to see whether actively selecting preference queries would lead to faster adaptation than passive human feedback. Remember that for \textbf{Ours (Active)} the robot asked questions using \eq{M7}, while for \textbf{Ours (Passive)} the robot chose preference queries at random. We observe that \textbf{Ours (Active)} has significantly better performance than \textbf{Ours (Passive)} across $500$ users ($p < 0.05$). We conclude that hypothesis \textbf{H9} is supported when robots learn from demonstrations and preferences.
\section{Conclusion}

In this paper we developed an alternate formalism for learning from physical human-robot interaction that unifies demonstrations, corrections, and preferences. When humans and robots share spaces, physical interaction is inevitable. Robots should leverage this interaction to learn from the human and improve their own behavior. But physical interaction takes many forms: humans can kinesthetically guide the robot throughout an entire task, fine-tune snippets of the robot's motion, or indicate which robot trajectories they prefer. Existing methods either learn from one of these interaction modalities, or combine multiple modalities by assuming prior information about the human's task. Instead, we introduce an end-to-end framework that (a) learns a reward function from scratch and then (b) optimizes this reward function to obtain robot trajectories.

Our key technical insight was that we can unite demonstrations, corrections, and preferences within the same framework by learning to assign higher rewards to these human inputs than to nearby alternatives. We first described a way to generate trajectory modifications. Next, we derived loss functions for learning from demonstrations, corrections, and preferences, and used these loss functions to train an ensemble of reward models. We also enabled robots to actively prompt the human and gain information about the correct task behavior. Finally, we converted the robot's learned rewards to robot trajectories using constrained optimization. Our framework was specifically developed for robot arms performing manipulation tasks: through simulations and a user study we compared our approach to multiple state-of-the-art baselines. Our results indicate that --- when the robot knows what tasks to expect --- our learning approach is comparable to existing methods that rely on pre-programmed features. However, when the robot encounters unexpected tasks (or when the robot must learn a new task from scratch), our method outperforms the interactive reward learning and imitation learning baselines.

\p{Limitations and Future Works} Our work is a step towards seamless communication between humans and robot arms. Because our system can learn new behaviors, one practical concern is \textit{safety}: we must ensure that the robot learns trajectories that are safe for shared human-robot spaces. For instance, in our running example the robot arm should never learn to swing towards the human or run into the table. If the designer knows these safety constraints \textit{a priori} we can embed them within our approach. Specifically, designers could augment \eq{M10} to constrain the robot to have a certain workspace or speed thresholds. However, if designers do not impose any limits, we currently cannot guarantee that our robot will learn human-friendly behaviors. 

One assumption throughout our work is that the human’s inputs are noisily optimal. We assume that — when the human makes a correction or provides a preference — on average their input is better aligned with their underlying reward function than the alternatives. However, in some settings and modalities the human’s inputs may have a persistent bias, leading to suboptimal demonstrations, corrections, or preferences. Imagine a person teaching a robot to move across the table: if this human teacher cannot reach the opposite side of the table, all of their demonstrations may only move the robot part of the way to their intended goal. Existing works have explored methods for extrapolating from suboptimal demonstrations to reach performance that exceeds the given feedback \cite{chen2021learning, brown2020better, schrum2022mind}. We hypothesize that these methods could be combined with our reward learning approach by leveraging the diverse types of human feedback. For example, the user may have a persistent bias in their demonstrations but not their preferences (e.g., in our example the human cannot reach across the table but they can compare two trajectories that do so). Hence, we envision an iterative solution where the robot uses our approach to build a reward model from different types of feedback while identifying which of these modalities are biased and which are reliable.

\section{Acknowledgements}

This work was funded in part by NSF Grant $\#2129201$.

\bibliographystyle{ACM-Reference-Format}
\bibliography{sample-bibliography}


\begin{thebibliography}{50}


\ifx \showCODEN    \undefined \def \showCODEN     #1{\unskip}     \fi
\ifx \showDOI      \undefined \def \showDOI       #1{#1}\fi
\ifx \showISBNx    \undefined \def \showISBNx     #1{\unskip}     \fi
\ifx \showISBNxiii \undefined \def \showISBNxiii  #1{\unskip}     \fi
\ifx \showISSN     \undefined \def \showISSN      #1{\unskip}     \fi
\ifx \showLCCN     \undefined \def \showLCCN      #1{\unskip}     \fi
\ifx \shownote     \undefined \def \shownote      #1{#1}          \fi
\ifx \showarticletitle \undefined \def \showarticletitle #1{#1}   \fi
\ifx \showURL      \undefined \def \showURL       {\relax}        \fi
\providecommand\bibfield[2]{#2}
\providecommand\bibinfo[2]{#2}
\providecommand\natexlab[1]{#1}
\providecommand\showeprint[2][]{arXiv:#2}

\bibitem[\protect\citeauthoryear{Abbeel and Ng}{Abbeel and Ng}{2004}]%
        {abbeel2004apprenticeship}
\bibfield{author}{\bibinfo{person}{Pieter Abbeel} {and} \bibinfo{person}{Andrew~Y Ng}.} \bibinfo{year}{2004}\natexlab{}.
\newblock \showarticletitle{Apprenticeship learning via inverse reinforcement learning}. In \bibinfo{booktitle}{\emph{International Conference on Machine Learning}}.
\newblock


\bibitem[\protect\citeauthoryear{Akgun, Cakmak, Jiang, and Thomaz}{Akgun et~al\mbox{.}}{2012}]%
        {akgun2012keyframe}
\bibfield{author}{\bibinfo{person}{Baris Akgun}, \bibinfo{person}{Maya Cakmak}, \bibinfo{person}{Karl Jiang}, {and} \bibinfo{person}{Andrea~L Thomaz}.} \bibinfo{year}{2012}\natexlab{}.
\newblock \showarticletitle{Keyframe-based learning from demonstration}.
\newblock \bibinfo{journal}{\emph{International Journal of Social Robotics}} \bibinfo{volume}{4}, \bibinfo{number}{4} (\bibinfo{year}{2012}), \bibinfo{pages}{343--355}.
\newblock


\bibitem[\protect\citeauthoryear{Argall, Chernova, Veloso, and Browning}{Argall et~al\mbox{.}}{2009}]%
        {argall2009survey}
\bibfield{author}{\bibinfo{person}{Brenna~D Argall}, \bibinfo{person}{Sonia Chernova}, \bibinfo{person}{Manuela Veloso}, {and} \bibinfo{person}{Brett Browning}.} \bibinfo{year}{2009}\natexlab{}.
\newblock \showarticletitle{A survey of robot learning from demonstration}.
\newblock \bibinfo{journal}{\emph{Robotics and Autonomous Systems}} \bibinfo{volume}{57}, \bibinfo{number}{5} (\bibinfo{year}{2009}), \bibinfo{pages}{469--483}.
\newblock


\bibitem[\protect\citeauthoryear{B{\i}y{\i}k, Losey, Palan, Landolfi, Shevchuk, and Sadigh}{B{\i}y{\i}k et~al\mbox{.}}{2021}]%
        {biyik2021learning}
\bibfield{author}{\bibinfo{person}{Erdem B{\i}y{\i}k}, \bibinfo{person}{Dylan~P Losey}, \bibinfo{person}{Malayandi Palan}, \bibinfo{person}{Nicholas~C Landolfi}, \bibinfo{person}{Gleb Shevchuk}, {and} \bibinfo{person}{Dorsa Sadigh}.} \bibinfo{year}{2021}\natexlab{}.
\newblock \showarticletitle{Learning reward functions from diverse sources of human feedback: {O}ptimally integrating demonstrations and preferences}.
\newblock \bibinfo{journal}{\emph{The International Journal of Robotics Research}} (\bibinfo{year}{2021}).
\newblock


\bibitem[\protect\citeauthoryear{Bobu, Bajcsy, Fisac, Deglurkar, and Dragan}{Bobu et~al\mbox{.}}{2020}]%
        {bobu2020quantifying}
\bibfield{author}{\bibinfo{person}{Andreea Bobu}, \bibinfo{person}{Andrea Bajcsy}, \bibinfo{person}{Jaime~F Fisac}, \bibinfo{person}{Sampada Deglurkar}, {and} \bibinfo{person}{Anca~D Dragan}.} \bibinfo{year}{2020}\natexlab{}.
\newblock \showarticletitle{Quantifying hypothesis space misspecification in learning from human--robot demonstrations and physical corrections}.
\newblock \bibinfo{journal}{\emph{IEEE Transactions on Robotics}} \bibinfo{volume}{36}, \bibinfo{number}{3} (\bibinfo{year}{2020}), \bibinfo{pages}{835--854}.
\newblock


\bibitem[\protect\citeauthoryear{Bobu, Wiggert, Tomlin, and Dragan}{Bobu et~al\mbox{.}}{2022}]%
        {bobu2022inducing}
\bibfield{author}{\bibinfo{person}{Andreea Bobu}, \bibinfo{person}{Marius Wiggert}, \bibinfo{person}{Claire Tomlin}, {and} \bibinfo{person}{Anca~D Dragan}.} \bibinfo{year}{2022}\natexlab{}.
\newblock \showarticletitle{Inducing Structure in Reward Learning by Learning Features}.
\newblock \bibinfo{journal}{\emph{The International Journal of Robotics Research}} (\bibinfo{year}{2022}).
\newblock


\bibitem[\protect\citeauthoryear{Brown, Goo, Nagarajan, and Niekum}{Brown et~al\mbox{.}}{2019}]%
        {brown2019extrapolating}
\bibfield{author}{\bibinfo{person}{Daniel Brown}, \bibinfo{person}{Wonjoon Goo}, \bibinfo{person}{Prabhat Nagarajan}, {and} \bibinfo{person}{Scott Niekum}.} \bibinfo{year}{2019}\natexlab{}.
\newblock \showarticletitle{Extrapolating beyond suboptimal demonstrations via inverse reinforcement learning from observations}. In \bibinfo{booktitle}{\emph{International Conference on Machine Learning}}. \bibinfo{pages}{783--792}.
\newblock


\bibitem[\protect\citeauthoryear{Brown, Goo, and Niekum}{Brown et~al\mbox{.}}{2020}]%
        {brown2020better}
\bibfield{author}{\bibinfo{person}{Daniel~S Brown}, \bibinfo{person}{Wonjoon Goo}, {and} \bibinfo{person}{Scott Niekum}.} \bibinfo{year}{2020}\natexlab{}.
\newblock \showarticletitle{Better-than-demonstrator imitation learning via automatically-ranked demonstrations}. In \bibinfo{booktitle}{\emph{Conference on robot learning}}. PMLR, \bibinfo{pages}{330--359}.
\newblock


\bibitem[\protect\citeauthoryear{Chen, Paleja, and Gombolay}{Chen et~al\mbox{.}}{2021}]%
        {chen2021learning}
\bibfield{author}{\bibinfo{person}{Letian Chen}, \bibinfo{person}{Rohan Paleja}, {and} \bibinfo{person}{Matthew Gombolay}.} \bibinfo{year}{2021}\natexlab{}.
\newblock \showarticletitle{Learning from suboptimal demonstration via self-supervised reward regression}. In \bibinfo{booktitle}{\emph{Conference on Robot Learning}}. \bibinfo{pages}{1262--1277}.
\newblock


\bibitem[\protect\citeauthoryear{Christiano, Leike, Brown, Martic, Legg, and Amodei}{Christiano et~al\mbox{.}}{2017}]%
        {christiano2017deep}
\bibfield{author}{\bibinfo{person}{Paul~F Christiano}, \bibinfo{person}{Jan Leike}, \bibinfo{person}{Tom Brown}, \bibinfo{person}{Miljan Martic}, \bibinfo{person}{Shane Legg}, {and} \bibinfo{person}{Dario Amodei}.} \bibinfo{year}{2017}\natexlab{}.
\newblock \showarticletitle{Deep reinforcement learning from human preferences}. In \bibinfo{booktitle}{\emph{Advances in Neural Information Processing Systems}}.
\newblock


\bibitem[\protect\citeauthoryear{Cover}{Cover}{1999}]%
        {cover1999elements}
\bibfield{author}{\bibinfo{person}{Thomas~M Cover}.} \bibinfo{year}{1999}\natexlab{}.
\newblock \bibinfo{booktitle}{\emph{Elements of Information Theory}}.
\newblock \bibinfo{publisher}{John Wiley \& Sons}.
\newblock


\bibitem[\protect\citeauthoryear{De~Santis, Siciliano, De~Luca, and Bicchi}{De~Santis et~al\mbox{.}}{2008}]%
        {de2008atlas}
\bibfield{author}{\bibinfo{person}{Agostino De~Santis}, \bibinfo{person}{Bruno Siciliano}, \bibinfo{person}{Alessandro De~Luca}, {and} \bibinfo{person}{Antonio Bicchi}.} \bibinfo{year}{2008}\natexlab{}.
\newblock \showarticletitle{An atlas of physical human--robot interaction}.
\newblock \bibinfo{journal}{\emph{Mechanism and Machine Theory}} \bibinfo{volume}{43}, \bibinfo{number}{3} (\bibinfo{year}{2008}), \bibinfo{pages}{253--270}.
\newblock


\bibitem[\protect\citeauthoryear{Dragan, Muelling, Bagnell, and Srinivasa}{Dragan et~al\mbox{.}}{2015}]%
        {dragan2015movement}
\bibfield{author}{\bibinfo{person}{Anca~D Dragan}, \bibinfo{person}{Katharina Muelling}, \bibinfo{person}{J~Andrew Bagnell}, {and} \bibinfo{person}{Siddhartha~S Srinivasa}.} \bibinfo{year}{2015}\natexlab{}.
\newblock \showarticletitle{Movement primitives via optimization}. In \bibinfo{booktitle}{\emph{IEEE International Conference on Robotics and Automation}}. \bibinfo{pages}{2339--2346}.
\newblock


\bibitem[\protect\citeauthoryear{Fu, Luo, and Levine}{Fu et~al\mbox{.}}{2018}]%
        {fu2017learning}
\bibfield{author}{\bibinfo{person}{Justin Fu}, \bibinfo{person}{Katie Luo}, {and} \bibinfo{person}{Sergey Levine}.} \bibinfo{year}{2018}\natexlab{}.
\newblock \showarticletitle{Learning robust rewards with adversarial inverse reinforcement learning}. In \bibinfo{booktitle}{\emph{International Conference on Learning Representations}}.
\newblock


\bibitem[\protect\citeauthoryear{Gill, Murray, and Saunders}{Gill et~al\mbox{.}}{2005}]%
        {gill2005snopt}
\bibfield{author}{\bibinfo{person}{Philip~E Gill}, \bibinfo{person}{Walter Murray}, {and} \bibinfo{person}{Michael~A Saunders}.} \bibinfo{year}{2005}\natexlab{}.
\newblock \showarticletitle{{SNOPT: An SQP} algorithm for large-scale constrained optimization}.
\newblock \bibinfo{journal}{\emph{SIAM review}} \bibinfo{volume}{47}, \bibinfo{number}{1} (\bibinfo{year}{2005}), \bibinfo{pages}{99--131}.
\newblock


\bibitem[\protect\citeauthoryear{Haddadin, Albu-Schaffer, De~Luca, and Hirzinger}{Haddadin et~al\mbox{.}}{2008}]%
        {haddadin2008collision}
\bibfield{author}{\bibinfo{person}{Sami Haddadin}, \bibinfo{person}{Alin Albu-Schaffer}, \bibinfo{person}{Alessandro De~Luca}, {and} \bibinfo{person}{Gerd Hirzinger}.} \bibinfo{year}{2008}\natexlab{}.
\newblock \showarticletitle{Collision detection and reaction: {A} contribution to safe physical human-robot interaction}. In \bibinfo{booktitle}{\emph{IEEE/RSJ International Conference on Intelligent Robots and Systems}}. \bibinfo{pages}{3356--3363}.
\newblock


\bibitem[\protect\citeauthoryear{Haddadin and Croft}{Haddadin and Croft}{2016}]%
        {haddadin2016physical}
\bibfield{author}{\bibinfo{person}{Sami Haddadin} {and} \bibinfo{person}{Elizabeth Croft}.} \bibinfo{year}{2016}\natexlab{}.
\newblock \showarticletitle{Physical human--robot interaction}.
\newblock In \bibinfo{booktitle}{\emph{Springer Handbook of Robotics}}. \bibinfo{pages}{1835--1874}.
\newblock


\bibitem[\protect\citeauthoryear{Hagenow, Senft, Radwin, Gleicher, Mutlu, and Zinn}{Hagenow et~al\mbox{.}}{2021}]%
        {hagenow2021corrective}
\bibfield{author}{\bibinfo{person}{Michael Hagenow}, \bibinfo{person}{Emmanuel Senft}, \bibinfo{person}{Robert Radwin}, \bibinfo{person}{Michael Gleicher}, \bibinfo{person}{Bilge Mutlu}, {and} \bibinfo{person}{Michael Zinn}.} \bibinfo{year}{2021}\natexlab{}.
\newblock \showarticletitle{Corrective shared autonomy for addressing task variability}.
\newblock \bibinfo{journal}{\emph{IEEE Robotics and Automation Letters}} \bibinfo{volume}{6}, \bibinfo{number}{2} (\bibinfo{year}{2021}), \bibinfo{pages}{3720--3727}.
\newblock


\bibitem[\protect\citeauthoryear{Hogan}{Hogan}{1984}]%
        {hogan1984impedance}
\bibfield{author}{\bibinfo{person}{Neville Hogan}.} \bibinfo{year}{1984}\natexlab{}.
\newblock \showarticletitle{Impedance control: {A}n approach to manipulation}. In \bibinfo{booktitle}{\emph{American Control Conference}}. \bibinfo{pages}{304--313}.
\newblock


\bibitem[\protect\citeauthoryear{Hoque, Balakrishna, Novoseller, Wilcox, Brown, and Goldberg}{Hoque et~al\mbox{.}}{2021}]%
        {hoque2021thriftydagger}
\bibfield{author}{\bibinfo{person}{Ryan Hoque}, \bibinfo{person}{Ashwin Balakrishna}, \bibinfo{person}{Ellen Novoseller}, \bibinfo{person}{Albert Wilcox}, \bibinfo{person}{Daniel~S Brown}, {and} \bibinfo{person}{Ken Goldberg}.} \bibinfo{year}{2021}\natexlab{}.
\newblock \showarticletitle{{ThriftyDAgger: B}udget-aware novelty and risk gating for interactive imitation learning}. In \bibinfo{booktitle}{\emph{Conference on Robot Learning}}.
\newblock


\bibitem[\protect\citeauthoryear{Howell, Jackson, and Manchester}{Howell et~al\mbox{.}}{2019}]%
        {howell2019altro}
\bibfield{author}{\bibinfo{person}{Taylor~A Howell}, \bibinfo{person}{Brian~E Jackson}, {and} \bibinfo{person}{Zachary Manchester}.} \bibinfo{year}{2019}\natexlab{}.
\newblock \showarticletitle{{ALTRO: A} fast solver for constrained trajectory optimization}. In \bibinfo{booktitle}{\emph{IEEE/RSJ International Conference on Intelligent Robots and Systems}}. \bibinfo{pages}{7674--7679}.
\newblock


\bibitem[\protect\citeauthoryear{Ibarz, Leike, Pohlen, Irving, Legg, and Amodei}{Ibarz et~al\mbox{.}}{2018}]%
        {ibarz2018reward}
\bibfield{author}{\bibinfo{person}{Borja Ibarz}, \bibinfo{person}{Jan Leike}, \bibinfo{person}{Tobias Pohlen}, \bibinfo{person}{Geoffrey Irving}, \bibinfo{person}{Shane Legg}, {and} \bibinfo{person}{Dario Amodei}.} \bibinfo{year}{2018}\natexlab{}.
\newblock \showarticletitle{Reward learning from human preferences and demonstrations in {Atari}}. In \bibinfo{booktitle}{\emph{Advances in Neural Information Processing Systems}}.
\newblock


\bibitem[\protect\citeauthoryear{Jain, Sharma, Joachims, and Saxena}{Jain et~al\mbox{.}}{2015}]%
        {jain2015learning}
\bibfield{author}{\bibinfo{person}{Ashesh Jain}, \bibinfo{person}{Shikhar Sharma}, \bibinfo{person}{Thorsten Joachims}, {and} \bibinfo{person}{Ashutosh Saxena}.} \bibinfo{year}{2015}\natexlab{}.
\newblock \showarticletitle{Learning preferences for manipulation tasks from online coactive feedback}.
\newblock \bibinfo{journal}{\emph{The International Journal of Robotics Research}} \bibinfo{volume}{34}, \bibinfo{number}{10} (\bibinfo{year}{2015}), \bibinfo{pages}{1296--1313}.
\newblock


\bibitem[\protect\citeauthoryear{Jeon, Milli, and Dragan}{Jeon et~al\mbox{.}}{2020}]%
        {jeon2020reward}
\bibfield{author}{\bibinfo{person}{Hong~Jun Jeon}, \bibinfo{person}{Smitha Milli}, {and} \bibinfo{person}{Anca Dragan}.} \bibinfo{year}{2020}\natexlab{}.
\newblock \showarticletitle{Reward-rational (implicit) choice: {A} unifying formalism for reward learning}. In \bibinfo{booktitle}{\emph{Advances in Neural Information Processing Systems}}. \bibinfo{pages}{4415--4426}.
\newblock


\bibitem[\protect\citeauthoryear{Kalakrishnan, Pastor, Righetti, and Schaal}{Kalakrishnan et~al\mbox{.}}{2013}]%
        {kalakrishnan2013learning}
\bibfield{author}{\bibinfo{person}{Mrinal Kalakrishnan}, \bibinfo{person}{Peter Pastor}, \bibinfo{person}{Ludovic Righetti}, {and} \bibinfo{person}{Stefan Schaal}.} \bibinfo{year}{2013}\natexlab{}.
\newblock \showarticletitle{Learning objective functions for manipulation}. In \bibinfo{booktitle}{\emph{IEEE International Conference on Robotics and Automation}}. \bibinfo{pages}{1331--1336}.
\newblock


\bibitem[\protect\citeauthoryear{Kelly, Sidrane, Driggs-Campbell, and Kochenderfer}{Kelly et~al\mbox{.}}{2019}]%
        {kelly2019hg}
\bibfield{author}{\bibinfo{person}{Michael Kelly}, \bibinfo{person}{Chelsea Sidrane}, \bibinfo{person}{Katherine Driggs-Campbell}, {and} \bibinfo{person}{Mykel~J Kochenderfer}.} \bibinfo{year}{2019}\natexlab{}.
\newblock \showarticletitle{{HG-DA}gger: {I}nteractive imitation learning with human experts}. In \bibinfo{booktitle}{\emph{International Conference on Robotics and Automation}}. \bibinfo{pages}{8077--8083}.
\newblock


\bibitem[\protect\citeauthoryear{Khoramshahi and Billard}{Khoramshahi and Billard}{2019}]%
        {khoramshahi2019dynamical}
\bibfield{author}{\bibinfo{person}{Mahdi Khoramshahi} {and} \bibinfo{person}{Aude Billard}.} \bibinfo{year}{2019}\natexlab{}.
\newblock \showarticletitle{A dynamical system approach to task-adaptation in physical human--robot interaction}.
\newblock \bibinfo{journal}{\emph{Autonomous Robots}} \bibinfo{volume}{43}, \bibinfo{number}{4} (\bibinfo{year}{2019}), \bibinfo{pages}{927--946}.
\newblock


\bibitem[\protect\citeauthoryear{Kingma and Ba}{Kingma and Ba}{2014}]%
        {kingma2014adam}
\bibfield{author}{\bibinfo{person}{Diederik~P Kingma} {and} \bibinfo{person}{Jimmy Ba}.} \bibinfo{year}{2014}\natexlab{}.
\newblock \showarticletitle{{Adam: A} method for stochastic optimization}.
\newblock \bibinfo{journal}{\emph{arXiv preprint arXiv:1412.6980}} (\bibinfo{year}{2014}).
\newblock


\bibitem[\protect\citeauthoryear{Lee, Smith, and Abbeel}{Lee et~al\mbox{.}}{2021}]%
        {lee2021pebble}
\bibfield{author}{\bibinfo{person}{Kimin Lee}, \bibinfo{person}{Laura~M Smith}, {and} \bibinfo{person}{Pieter Abbeel}.} \bibinfo{year}{2021}\natexlab{}.
\newblock \showarticletitle{{PEBBLE: F}eedback-efficient interactive reinforcement learning via relabeling experience and unsupervised pre-training}. In \bibinfo{booktitle}{\emph{International Conference on Machine Learning}}. \bibinfo{pages}{6152--6163}.
\newblock


\bibitem[\protect\citeauthoryear{Li, Canberk, Losey, and Sadigh}{Li et~al\mbox{.}}{2021}]%
        {li2021learning}
\bibfield{author}{\bibinfo{person}{Mengxi Li}, \bibinfo{person}{Alper Canberk}, \bibinfo{person}{Dylan~P Losey}, {and} \bibinfo{person}{Dorsa Sadigh}.} \bibinfo{year}{2021}\natexlab{}.
\newblock \showarticletitle{Learning human objectives from sequences of physical corrections}. In \bibinfo{booktitle}{\emph{IEEE International Conference on Robotics and Automation}}. \bibinfo{pages}{2877--2883}.
\newblock


\bibitem[\protect\citeauthoryear{Li, Carboni, Gonzalez, Campolo, and Burdet}{Li et~al\mbox{.}}{2019}]%
        {li2019differential}
\bibfield{author}{\bibinfo{person}{Yanan Li}, \bibinfo{person}{Gerolamo Carboni}, \bibinfo{person}{Franck Gonzalez}, \bibinfo{person}{Domenico Campolo}, {and} \bibinfo{person}{Etienne Burdet}.} \bibinfo{year}{2019}\natexlab{}.
\newblock \showarticletitle{Differential game theory for versatile physical human--robot interaction}.
\newblock \bibinfo{journal}{\emph{Nature Machine Intelligence}} \bibinfo{volume}{1}, \bibinfo{number}{1} (\bibinfo{year}{2019}), \bibinfo{pages}{36--43}.
\newblock


\bibitem[\protect\citeauthoryear{Losey, Bajcsy, O’Malley, and Dragan}{Losey et~al\mbox{.}}{2021}]%
        {losey2021physical}
\bibfield{author}{\bibinfo{person}{Dylan~P Losey}, \bibinfo{person}{Andrea Bajcsy}, \bibinfo{person}{Marcia~K O’Malley}, {and} \bibinfo{person}{Anca~D Dragan}.} \bibinfo{year}{2021}\natexlab{}.
\newblock \showarticletitle{Physical interaction as communication: {L}earning robot objectives online from human corrections}.
\newblock \bibinfo{journal}{\emph{The International Journal of Robotics Research}} (\bibinfo{year}{2021}).
\newblock


\bibitem[\protect\citeauthoryear{Losey, McDonald, Battaglia, and O'Malley}{Losey et~al\mbox{.}}{2018}]%
        {losey2018review}
\bibfield{author}{\bibinfo{person}{Dylan~P Losey}, \bibinfo{person}{Craig~G McDonald}, \bibinfo{person}{Edoardo Battaglia}, {and} \bibinfo{person}{Marcia~K O'Malley}.} \bibinfo{year}{2018}\natexlab{}.
\newblock \showarticletitle{A review of intent detection, arbitration, and communication aspects of shared control for physical human--robot interaction}.
\newblock \bibinfo{journal}{\emph{Applied Mechanics Reviews}} \bibinfo{volume}{70}, \bibinfo{number}{1} (\bibinfo{year}{2018}).
\newblock


\bibitem[\protect\citeauthoryear{Losey and O'Malley}{Losey and O'Malley}{2017}]%
        {losey2017trajectory}
\bibfield{author}{\bibinfo{person}{Dylan~P Losey} {and} \bibinfo{person}{Marcia~K O'Malley}.} \bibinfo{year}{2017}\natexlab{}.
\newblock \showarticletitle{Trajectory deformations from physical human--robot interaction}.
\newblock \bibinfo{journal}{\emph{IEEE Transactions on Robotics}} \bibinfo{volume}{34}, \bibinfo{number}{1} (\bibinfo{year}{2017}), \bibinfo{pages}{126--138}.
\newblock


\bibitem[\protect\citeauthoryear{Losey and O'Malley}{Losey and O'Malley}{2019}]%
        {losey2019learning}
\bibfield{author}{\bibinfo{person}{Dylan~P Losey} {and} \bibinfo{person}{Marcia~K O'Malley}.} \bibinfo{year}{2019}\natexlab{}.
\newblock \showarticletitle{Learning the correct robot trajectory in real-time from physical human interactions}.
\newblock \bibinfo{journal}{\emph{ACM Transactions on Human-Robot Interaction}} \bibinfo{volume}{9}, \bibinfo{number}{1} (\bibinfo{year}{2019}), \bibinfo{pages}{1--19}.
\newblock


\bibitem[\protect\citeauthoryear{Lucas, Griffiths, Xu, and Fawcett}{Lucas et~al\mbox{.}}{2008}]%
        {lucas2008rational}
\bibfield{author}{\bibinfo{person}{Christopher Lucas}, \bibinfo{person}{Thomas Griffiths}, \bibinfo{person}{Fei Xu}, {and} \bibinfo{person}{Christine Fawcett}.} \bibinfo{year}{2008}\natexlab{}.
\newblock \showarticletitle{A rational model of preference learning and choice prediction by children}. In \bibinfo{booktitle}{\emph{Advances in Neural Information Processing Systems}}.
\newblock


\bibitem[\protect\citeauthoryear{Luce}{Luce}{2012}]%
        {luce2012individual}
\bibfield{author}{\bibinfo{person}{R~Duncan Luce}.} \bibinfo{year}{2012}\natexlab{}.
\newblock \bibinfo{booktitle}{\emph{Individual Choice Behavior: {A} Theoretical Analysis}}.
\newblock \bibinfo{publisher}{Courier Corporation}.
\newblock


\bibitem[\protect\citeauthoryear{M{\"o}rtl, Lawitzky, Kucukyilmaz, Sezgin, Basdogan, and Hirche}{M{\"o}rtl et~al\mbox{.}}{2012}]%
        {mortl2012role}
\bibfield{author}{\bibinfo{person}{Alexander M{\"o}rtl}, \bibinfo{person}{Martin Lawitzky}, \bibinfo{person}{Ayse Kucukyilmaz}, \bibinfo{person}{Metin Sezgin}, \bibinfo{person}{Cagatay Basdogan}, {and} \bibinfo{person}{Sandra Hirche}.} \bibinfo{year}{2012}\natexlab{}.
\newblock \showarticletitle{The role of roles: {P}hysical cooperation between humans and robots}.
\newblock \bibinfo{journal}{\emph{The International Journal of Robotics Research}} \bibinfo{volume}{31}, \bibinfo{number}{13} (\bibinfo{year}{2012}), \bibinfo{pages}{1656--1674}.
\newblock


\bibitem[\protect\citeauthoryear{Musi{\'c} and Hirche}{Musi{\'c} and Hirche}{2017}]%
        {music2017control}
\bibfield{author}{\bibinfo{person}{Selma Musi{\'c}} {and} \bibinfo{person}{Sandra Hirche}.} \bibinfo{year}{2017}\natexlab{}.
\newblock \showarticletitle{Control sharing in human-robot team interaction}.
\newblock \bibinfo{journal}{\emph{Annual Reviews in Control}}  \bibinfo{volume}{44} (\bibinfo{year}{2017}), \bibinfo{pages}{342--354}.
\newblock


\bibitem[\protect\citeauthoryear{Ng and Russell}{Ng and Russell}{2000}]%
        {ng2000algorithms}
\bibfield{author}{\bibinfo{person}{Andrew~Y Ng} {and} \bibinfo{person}{Stuart~J Russell}.} \bibinfo{year}{2000}\natexlab{}.
\newblock \showarticletitle{Algorithms for inverse reinforcement learning}. In \bibinfo{booktitle}{\emph{International Conference on Machine Learning}}.
\newblock


\bibitem[\protect\citeauthoryear{Osa, Pajarinen, Neumann, Bagnell, Abbeel, Peters, et~al\mbox{.}}{Osa et~al\mbox{.}}{2018}]%
        {osa2018algorithmic}
\bibfield{author}{\bibinfo{person}{Takayuki Osa}, \bibinfo{person}{Joni Pajarinen}, \bibinfo{person}{Gerhard Neumann}, \bibinfo{person}{J~Andrew Bagnell}, \bibinfo{person}{Pieter Abbeel}, \bibinfo{person}{Jan Peters}, {et~al\mbox{.}}} \bibinfo{year}{2018}\natexlab{}.
\newblock \showarticletitle{An algorithmic perspective on imitation learning}.
\newblock \bibinfo{journal}{\emph{Foundations and Trends in Robotics}} \bibinfo{volume}{7}, \bibinfo{number}{1-2} (\bibinfo{year}{2018}), \bibinfo{pages}{1--179}.
\newblock


\bibitem[\protect\citeauthoryear{Ross, Gordon, and Bagnell}{Ross et~al\mbox{.}}{2011}]%
        {ross2011reduction}
\bibfield{author}{\bibinfo{person}{St{\'e}phane Ross}, \bibinfo{person}{Geoffrey Gordon}, {and} \bibinfo{person}{Drew Bagnell}.} \bibinfo{year}{2011}\natexlab{}.
\newblock \showarticletitle{A reduction of imitation learning and structured prediction to no-regret online learning}. In \bibinfo{booktitle}{\emph{International Conference on Artificial Intelligence and Statistics}}. \bibinfo{pages}{627--635}.
\newblock


\bibitem[\protect\citeauthoryear{Schrum, Hedlund-Botti, Moorman, and Gombolay}{Schrum et~al\mbox{.}}{2022}]%
        {schrum2022mind}
\bibfield{author}{\bibinfo{person}{Mariah~L Schrum}, \bibinfo{person}{Erin Hedlund-Botti}, \bibinfo{person}{Nina Moorman}, {and} \bibinfo{person}{Matthew~C Gombolay}.} \bibinfo{year}{2022}\natexlab{}.
\newblock \showarticletitle{Mind meld: Personalized meta-learning for robot-centric imitation learning}. In \bibinfo{booktitle}{\emph{2022 17th ACM/IEEE International Conference on Human-Robot Interaction (HRI)}}. IEEE, \bibinfo{pages}{157--165}.
\newblock


\bibitem[\protect\citeauthoryear{Schrum, Johnson, Ghuy, and Gombolay}{Schrum et~al\mbox{.}}{2020}]%
        {schrum2020four}
\bibfield{author}{\bibinfo{person}{Mariah~L Schrum}, \bibinfo{person}{Michael Johnson}, \bibinfo{person}{Muyleng Ghuy}, {and} \bibinfo{person}{Matthew~C Gombolay}.} \bibinfo{year}{2020}\natexlab{}.
\newblock \showarticletitle{Four years in review: Statistical practices of likert scales in human-robot interaction studies}. In \bibinfo{booktitle}{\emph{Companion of the 2020 ACM/IEEE International Conference on Human-Robot Interaction}}. \bibinfo{pages}{43--52}.
\newblock


\bibitem[\protect\citeauthoryear{Schulman, Duan, Ho, Lee, Awwal, Bradlow, Pan, Patil, Goldberg, and Abbeel}{Schulman et~al\mbox{.}}{2014}]%
        {schulman2014motion}
\bibfield{author}{\bibinfo{person}{John Schulman}, \bibinfo{person}{Yan Duan}, \bibinfo{person}{Jonathan Ho}, \bibinfo{person}{Alex Lee}, \bibinfo{person}{Ibrahim Awwal}, \bibinfo{person}{Henry Bradlow}, \bibinfo{person}{Jia Pan}, \bibinfo{person}{Sachin Patil}, \bibinfo{person}{Ken Goldberg}, {and} \bibinfo{person}{Pieter Abbeel}.} \bibinfo{year}{2014}\natexlab{}.
\newblock \showarticletitle{Motion planning with sequential convex optimization and convex collision checking}.
\newblock \bibinfo{journal}{\emph{The International Journal of Robotics Research}} \bibinfo{volume}{33}, \bibinfo{number}{9} (\bibinfo{year}{2014}), \bibinfo{pages}{1251--1270}.
\newblock


\bibitem[\protect\citeauthoryear{Shepard}{Shepard}{1957}]%
        {shepard1957stimulus}
\bibfield{author}{\bibinfo{person}{Roger~N Shepard}.} \bibinfo{year}{1957}\natexlab{}.
\newblock \showarticletitle{Stimulus and response generalization: {A} stochastic model relating generalization to distance in psychological space}.
\newblock \bibinfo{journal}{\emph{Psychometrika}} \bibinfo{volume}{22}, \bibinfo{number}{4} (\bibinfo{year}{1957}), \bibinfo{pages}{325--345}.
\newblock


\bibitem[\protect\citeauthoryear{Spencer, Choudhury, Barnes, Schmittle, Chiang, Ramadge, and Srinivasa}{Spencer et~al\mbox{.}}{2022}]%
        {spencer2022expert}
\bibfield{author}{\bibinfo{person}{Jonathan Spencer}, \bibinfo{person}{Sanjiban Choudhury}, \bibinfo{person}{Matthew Barnes}, \bibinfo{person}{Matthew Schmittle}, \bibinfo{person}{Mung Chiang}, \bibinfo{person}{Peter Ramadge}, {and} \bibinfo{person}{Sidd Srinivasa}.} \bibinfo{year}{2022}\natexlab{}.
\newblock \showarticletitle{Expert intervention learning}.
\newblock \bibinfo{journal}{\emph{Autonomous Robots}} \bibinfo{volume}{46}, \bibinfo{number}{1} (\bibinfo{year}{2022}), \bibinfo{pages}{99--113}.
\newblock


\bibitem[\protect\citeauthoryear{Yin, Melo, Paiva, and Billard}{Yin et~al\mbox{.}}{2019}]%
        {yin2019ensemble}
\bibfield{author}{\bibinfo{person}{Hang Yin}, \bibinfo{person}{Francisco~S Melo}, \bibinfo{person}{Ana Paiva}, {and} \bibinfo{person}{Aude Billard}.} \bibinfo{year}{2019}\natexlab{}.
\newblock \showarticletitle{An ensemble inverse optimal control approach for robotic task learning and adaptation}.
\newblock \bibinfo{journal}{\emph{Autonomous Robots}} \bibinfo{volume}{43}, \bibinfo{number}{4} (\bibinfo{year}{2019}), \bibinfo{pages}{875--896}.
\newblock


\bibitem[\protect\citeauthoryear{Zhang, Cao, Sadigh, and Sui}{Zhang et~al\mbox{.}}{2021}]%
        {zhang2021confidence}
\bibfield{author}{\bibinfo{person}{Songyuan Zhang}, \bibinfo{person}{Zhangjie Cao}, \bibinfo{person}{Dorsa Sadigh}, {and} \bibinfo{person}{Yanan Sui}.} \bibinfo{year}{2021}\natexlab{}.
\newblock \showarticletitle{Confidence-aware imitation learning from demonstrations with varying optimality}. In \bibinfo{booktitle}{\emph{Advances in Neural Information Processing Systems}}.
\newblock


\bibitem[\protect\citeauthoryear{Ziebart, Maas, Bagnell, and Dey}{Ziebart et~al\mbox{.}}{2008}]%
        {ziebart2008maximum}
\bibfield{author}{\bibinfo{person}{Brian~D Ziebart}, \bibinfo{person}{Andrew~L Maas}, \bibinfo{person}{J~Andrew Bagnell}, {and} \bibinfo{person}{Anind~K Dey}.} \bibinfo{year}{2008}\natexlab{}.
\newblock \showarticletitle{Maximum entropy inverse reinforcement learning}. In \bibinfo{booktitle}{\emph{AAAI}}. \bibinfo{pages}{1433--1438}.
\newblock


\end{thebibliography}

\end{document}